\documentclass[sigconf,nonacm]{acmart}

\usepackage{caption}
\usepackage{subcaption}
\usepackage{fontenc}
\usepackage[utf8]{inputenc} 
\usepackage{hyperref}       
\usepackage{url}            
\usepackage{booktabs}       
\usepackage{amsfonts}       
\usepackage{nicefrac}       
\usepackage{microtype}      
\usepackage{lipsum}
\usepackage{graphicx}
\graphicspath{ {./images/} }
\usepackage{amsmath}
\usepackage{xcolor}
\usepackage{multirow}
\usepackage{algorithm}
\usepackage{algorithmic}
\usepackage{tabularx}
\usepackage{bm}
\acmConference[Preprint]{...}{...}{...}
\acmISBN{}
\acmDOI{}
\acmYear{}
\copyrightyear{}
\begin{document}
\title{From Lab to Wrist: Bridging Metabolic Monitoring and Consumer Wearables for Heart Rate and Oxygen Consumption Modeling}

\author{Barak Gahtan}
\affiliation{%
  \institution{Technion Israel Institute of Technology }
  \city{Haifa}
  \country{Israel}}
\email{barakgahtan@cs.technion.ac.il}
\author{Sanketh Vedula}
\affiliation{%
  \institution{Technion Israel Institute of Technology }
  \city{Haifa}
  \country{Israel}}
\email{sanketh@cs.technion.ac.il}
\author{Gil Samuelly Leichtag}
\affiliation{%
  \institution{University of Haifa}
  \city{Haifa}
  \country{Israel}}
\email{gilsamuelly@gmail.com}
\author{Einat Kodesh}
\affiliation{%
  \institution{University of Haifa}
  \city{Haifa}
  \country{Israel}}
\email{ekodesh@univ.haifa.ac.il}
\author{Alex M. Bronstein}
\affiliation{%
  \institution{Technion Israel Institute of Technology }
  \city{Haifa}
  \country{Israel}}
\email{bron@cs.technion.ac.il}
\renewcommand{\shortauthors}{Gahtan et al.}
\renewcommand{\shorttitle}{Lab to Wrist.}
\begin{abstract}
Understanding physiological responses during running is critical for performance optimization, tailored training prescriptions, and athlete health management. We introduce a comprehensive framework---what we believe to be the first capable of predicting instantaneous oxygen consumption (VO$_{2}$) trajectories exclusively from consumer-grade wearable data. Our approach employs two complementary physiological models: (1) accurate modeling of heart rate (HR) dynamics via a physiologically constrained ordinary differential equation (ODE) and neural Kalman filter, trained on over 3 million HR observations, achieving 1-second interval predictions with mean absolute errors as low as 2.81\,bpm (correlation 0.87); and (2) leveraging the principles of precise HR modeling, a novel VO$_{2}$ prediction architecture requiring only the initial second of VO$_{2}$ data for calibration, enabling robust, sequence-to-sequence metabolic demand estimation. Despite relying solely on smartwatch and chest-strap data, our method achieves mean absolute percentage errors of approximately 13\%, effectively capturing rapid physiological transitions and steady-state conditions across diverse running intensities. Our synchronized dataset, complemented by blood lactate measurements, further lays the foundation for future noninvasive metabolic zone identification. By embedding physiological constraints within modern machine learning, this framework democratizes advanced metabolic monitoring, bridging laboratory-grade accuracy and everyday accessibility, thus empowering both elite athletes and recreational fitness enthusiasts.
\end{abstract}

\ccsdesc[500]{Computing methodologies~Machine learning}
\ccsdesc[500]{Computing methodologies~Neural networks}
\ccsdesc[300]{Human-centered computing~Human computer interaction (HCI)}
\ccsdesc[500]{Applied computing~Health informatics}
\ccsdesc[500]{Applied computing~Consumer health}

\keywords{Modeling Physiological Responses during Exercise, Machine Learning, Oxygen Consumption Prediction, Heart Rate Prediction, Multi-Modal Dataset, Neural Kalman Filter, Neural ODE, Instantaneous VO2 Prediction}
\maketitle

\section{Introduction}\label{intro}
Heart rate (HR) and oxygen consumption (VO$_2$) jointly reveal cardiovascular performance and metabolic demand, making them essential for optimizing athletic performance, preventing overtraining, and safeguarding health for both elite athletes and recreational fitness enthusiasts~\cite{noakes2008hill,wiecha2023vo2max,mohajan2023longterm}. While HR is readily accessible via consumer-grade wearables, VO$_2$ measurement remains confined to specialized laboratory equipment (costing over \$30\,000) and expert supervision~\cite{strasser2018survival}, limiting broad access to critical metabolic insights.

To bridge this gap, we propose two complementary physiological models that rely exclusively on data from consumer-grade wearables. First, we introduce an advanced HR dynamics model based on neural ordinary differential equations (ODEs)~\citep{chen2018neural} and with neural Kalman filtering~\citep{maybeck1982stochastic} to capture cardiac responses to exercise intensity. This approach precisely infers HR from external parameters, compensates for photoplethysmography (PPG) signal loss, and reproduces cardiac behavior during varied exercise efforts. Second, and more significantly, we present, to the best of our knowledge, the first approach for direct estimation of instantaneous VO$_2$ trajectories from smartwatch data using a Kalman-inspired deep learning architecture~\citep{krishnan2015deep}. By requiring only the initial second of VO$_2$ data for calibration, our framework enables sophisticated metabolic monitoring without specialized laboratory equipment, a capability previously unavailable through consumer devices.

Our framework advances physiological monitoring in several key ways. We achieve a mean absolute error (MAE) of 2.81\,bpm (correlation 0.87) in 1-second HR predictions across diverse running conditions, demonstrating robustness to wearable signal dropouts. Building on this foundation, our VO$_2$ model generates complete metabolic trajectories with a mean absolute percentage error (MAPE) of approximately 13\%, accurately capturing both rapid transitions and steady-state conditions. We validate these results through leave-one-runner-out cross-validation against gold-standard portable Cosmed K5 measurements, showing strong generalization across individuals. To achieve this, we collected a first-of-its-kind rich synchronized dataset that simultaneously acquires data from consumer smartwatch (Garmin 965), chest-strap HR, portable Cosmed K5 metabolic system, video capture every 200m, and blood lactate measurements, which also lays the groundwork for future noninvasive metabolic zone classification.

To support HR modeling, we leveraged a comprehensive dataset of 831 running sessions from 20 participants (Section~\ref{RUNDATASET}), representing over 52,825 minutes (approximately 880 hours) of activity and 10,222.90km of cumulative distance; HR models were evaluated via leave-three-runner-out cross-validation against concurrently recorded smartwatch and chest-strap data. For VO$_2$ prediction, we conducted an IRB-approved clinical trial with ten highly trained runners, each completing two structured track sessions; VO$_2$ models were validated using leave-one-runner-out cross-validation against gold-standard Cosmed K5 measurements. All device streams including Garmin 965, chest HR strap, portable metabolic system, video captures every 200m, and blood lactate samples—were time-synchronized to ensure robust multimodal evaluation across individuals and exercise intensities.

By embedding physiology-informed constraints within modern machine learning (ML), our approach extracts laboratory-quality insights from everyday wearables. The generalizable principles emerging from this work include the importance of synchronized multimodal data collection for robust cross-modality modeling and a transferable architecture that democratizes advanced metabolic monitoring on consumer devices.

The rest of this paper is organized as follows: Section~\ref{RW} reviews related work; Section~\ref{dataset} describes our data and protocol; Section~\ref{Framework1} details HR dynamics modeling; Section~\ref{Framework2} presents VO$_2$ prediction; and Section~\ref{conclusion} concludes with limitations and future directions.
\section{Related Work}\label{RW}
The estimation of physiological parameters during exercise spans multiple disciplines, including exercise physiology and computer science. We review relevant literature across three interconnected domains—physiological modeling, wearable technology, and ML, before highlighting our distinct contributions.

\noindent\textbf{Physiological Modeling of Exercise Response}
Mechanistic models of VO$_2$ kinetics date back to Hill and Lupton’s foundational work relating exercise intensity to oxygen uptake and energy expenditure~\cite{noakes2008hill}. These principles were extended into compartmental gas‐exchange models~\cite{millet2023vo2max} and the classic two‐component framework distinguishing fast and slow VO$_2$ responses~\cite{paterson1991oxygen}, with later refinements incorporating time delays~\cite{cooper2022geometric} and intensity‐dependent parameters~\cite{barstow1991linear}. Recent studies emphasize inter‐individual variability in VO$_2$ kinetics across age, training status, and health conditions~\cite{grey2015age,hecksteden2018individual,murias2015aging}, but remain constrained by controlled lab protocols that fail to capture real‐world variability in intensity, environment, and motivation~\cite{argent2021importance,piil2018aging}. VO$_2$ responses also differ markedly between laboratory and free‐living contexts~\cite{ferreira2016oxygen}, influenced by altitude, humidity, fatigue, nutrition, and emotional state~\cite{aylwin2023thermoregulatory,james2024mixed}. Although tools like \texttt{VO$_2$FITTING} aim to streamline kinetics analysis~\cite{zacca2019vo2fitting}, high‐precision models still rely on expensive stationary gas analyzers.

\noindent\textbf{Wearable Technology and Exercise Monitoring}
Consumer wearables now provide continuous PPG based HR and inertial sensing outside the laboratory~\cite{spathis2022longitudinal,chow2020accuracy}. Although recent devices have improved HR measurement accuracy~\cite{reece2021assessing}, motion artifacts and signal dropouts remain significant challenges during high-intensity activities~\cite{sto2020measurement,nelson2019accuracy}. Many commercial platforms include proprietary VO$_2$max estimators, yet these algorithms often lack rigorous validation across diverse populations and exercise conditions~\cite{molina2022validity,mccormick2018predictability}. To bridge the gap between consumer-grade sensor outputs and laboratory-grade cardiorespiratory metrics, cross-modal inference techniques have been proposed to translate wearable data into VO$_2$ and other physiological parameters~\cite{badrinarayanan2024evaluating,carrier2023validation}, but they frequently rely on steady-state assumptions or are evaluated on narrow cohorts. Emerging integrations of SpO$_2$ and ECG tracking show promise for reducing estimation errors~\cite{etiwy2019accuracy}, yet advanced algorithmic approaches remain essential to overcome the inherent limitations of wearable hardware in real-world exercise scenarios.

\noindent\textbf{ML for Physiological Parameter Estimation}
Recent years have witnessed a surge in applying ML techniques to physiological data analysis. Traditional approaches used statistical models to relate HR to VO$_{2}$~\cite{safari2022review}, but these models typically assume steady-state conditions and struggle with dynamic exercise.

Deep learning approaches have shown promise in processing multivariate physiological signals~\cite{hedge2021deep}, with recurrent and convolutional architectures demonstrating particular efficacy for temporal data. Attention mechanisms further boost performance by focusing on relevant signal patterns across various time scales~\cite{khurshid2023deep}, enabling systems to better detect meaningful physiological events and limit false alarms.

Several recent studies have applied hybrid modeling to HR dynamics during exercise~\cite{gentilin2023informative,nazaret2023modeling}, demonstrating improved accuracy compared to purely mechanistic or purely data-driven approaches. However, the extension of these methods to VO$_{2}$ modeling from wearable data remains largely unexplored, particularly in settings involving variable-intensity exercise and diverse subject populations~\citep{schumacher2024development}. \textbf{Notably absent from the literature is any method capable of predicting instantaneous VO$_{2}$ trajectories using only consumer wearables—a gap our work directly addresses.}

\noindent\textbf{Contributions and Distinctions from Prior Work.}
Our research addresses these gaps through two major contributions. First, we advance HR prediction during exercise with novel modeling approaches that differ from recent systems in several significant ways. Second, we introduce the first sequence-to-sequence approach for VO$_{2}$ estimation that uses only consumer-wearable data, bridging laboratory precision and real-world accessibility.

Our HR prediction model delivers 1-second temporal predictions, improving upon the 10-second averaged measurements of Nazaret et al.~\cite{nazaret2023modeling}. This granularity captures rapid cardiovascular transitions during variable-intensity exercise, enabling responsive workout feedback. Our system offers greater personalization flexibility. Unlike Nazaret et al., which requires extensive historical data per user, our approach generalizes effectively for new users with no historical data—critical for deployments where prior data is unavailable. This is achieved via our neural ODE framework or Kalman filter architecture, capturing physiological responses while adapting to individual cardiovascular characteristics without extensive calibration.

We enhance the practical applicability of our HR modeling through dual evaluation protocols that reflect realistic usage scenarios. Specifically, we utilize non-overlapping windows to evaluate our model under two distinct conditions: (1)~a continuous monitoring condition where the initial HR measurement is known for each window, and (2)~a minimal-data condition where only the first second's measurement (the first sample of the first window) is available for the entire exercise session (up to 120 minutes). This second protocol represents a significant challenge as it requires the model to maintain accurate predictions over extended durations with extremely limited initialization data. This approach transforms our sequence-to-sequence prediction into a generative forecasting model, demonstrating robust long-term predictive capability with minimal initialization data, albeit with an expected accuracy trade-off that we quantify and analyze.

These improvements collectively advance HR prediction from wearable data during real-world exercise conditions, with particular strengths in handling variable-intensity workouts across diverse populations, enabling responsive, personalized interfaces for exercise guidance without requiring extensive user-specific calibration periods.

Building on our HR modeling advances, we introduce the first sequence-to-sequence modeling approach for VO$_{2}$ estimation using only data available from consumer wearables. Unlike previous approaches that either (1) rely on steady-state assumptions~\cite{safari2022review} which fail during variable intensity exercise, or (2) require specialized equipment for direct measurement~\cite{strasser2018survival}, our method captures the dynamic nature of VO$_{2}$ during real-world variable-intensity exercise using only consumer-grade devices. Our VO$_{2}$ estimation method employs a Kalman filter architecture that learns shared physiological parameters among runners. Our approach demonstrates significant improvements over baseline methods, achieving mean absolute percentage errors of approximately 13\% for most participants across diverse running intensities. The model explicitly accounts for individual differences in physiological parameters through a learnable state-space representation, allowing personalization while maintaining the interpretability necessary for sports science applications. 

Together, these advances in HR dynamics modeling and VO$_{2}$ estimation form an integrated system that bridges the gap between laboratory-grade physiological assessment and consumer-accessible wearable technology, representing a significant step toward democratizing advanced exercise science for both research and practical applications.

\section{Experimental Protocol and Data Collection}
\begin{figure}[t]
\vspace{-12pt} 
\centering
\caption{Runners wearing \textit{Garmin 965} smartwatches, chest strap and Cosmed K5 portable metabolic systems during experimental sessions on an athletic track}
\captionsetup{justification=centering}
\begin{subfigure}{.24\columnwidth}
  \centering
  \captionsetup{justification=centering}
  \includegraphics[width=1\textwidth]{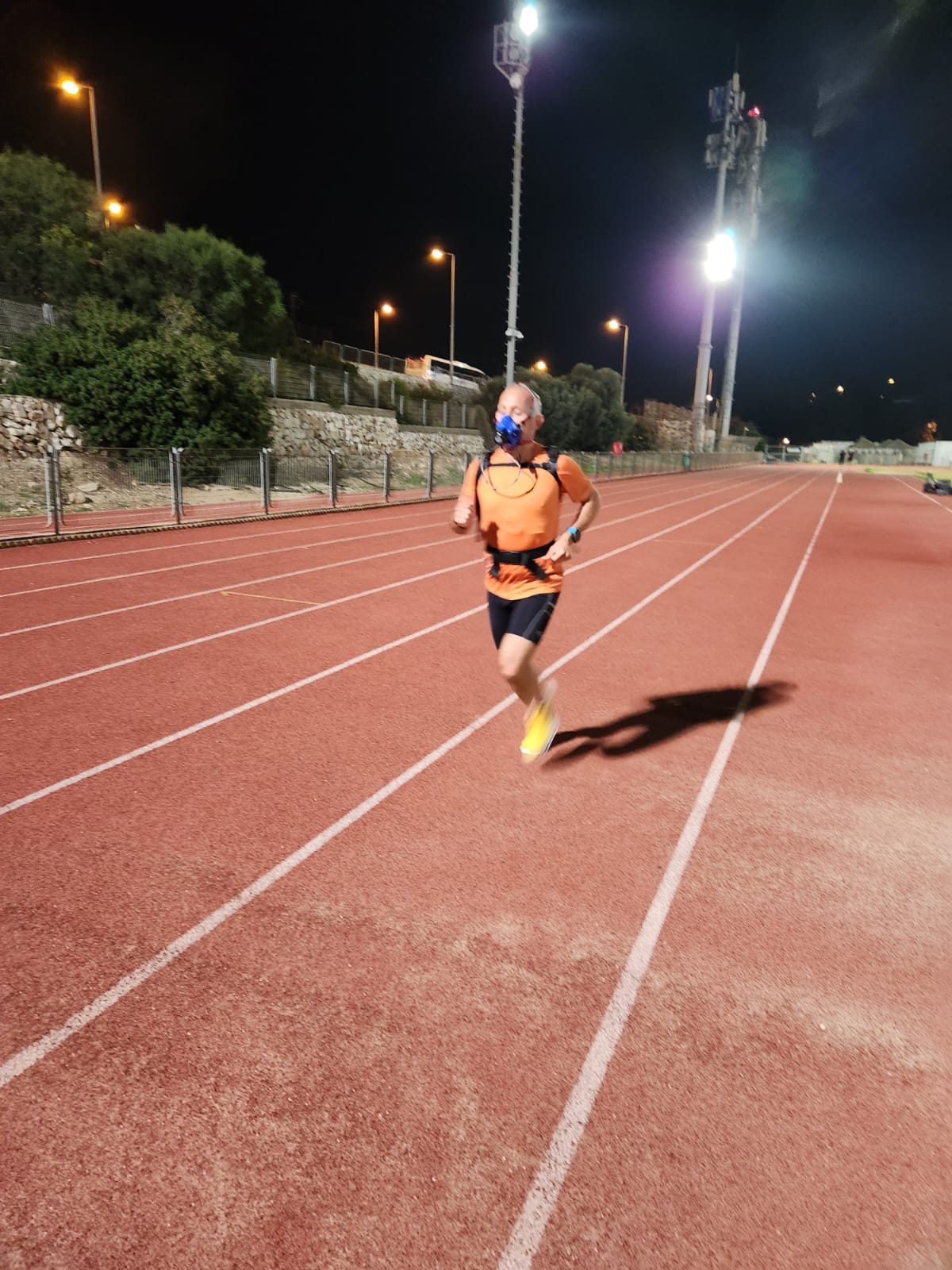}
\end{subfigure}%
\hspace{0.01\columnwidth}%
\begin{subfigure}{.24\columnwidth}
  \centering
  \captionsetup{justification=centering}
  \includegraphics[width=1\textwidth]{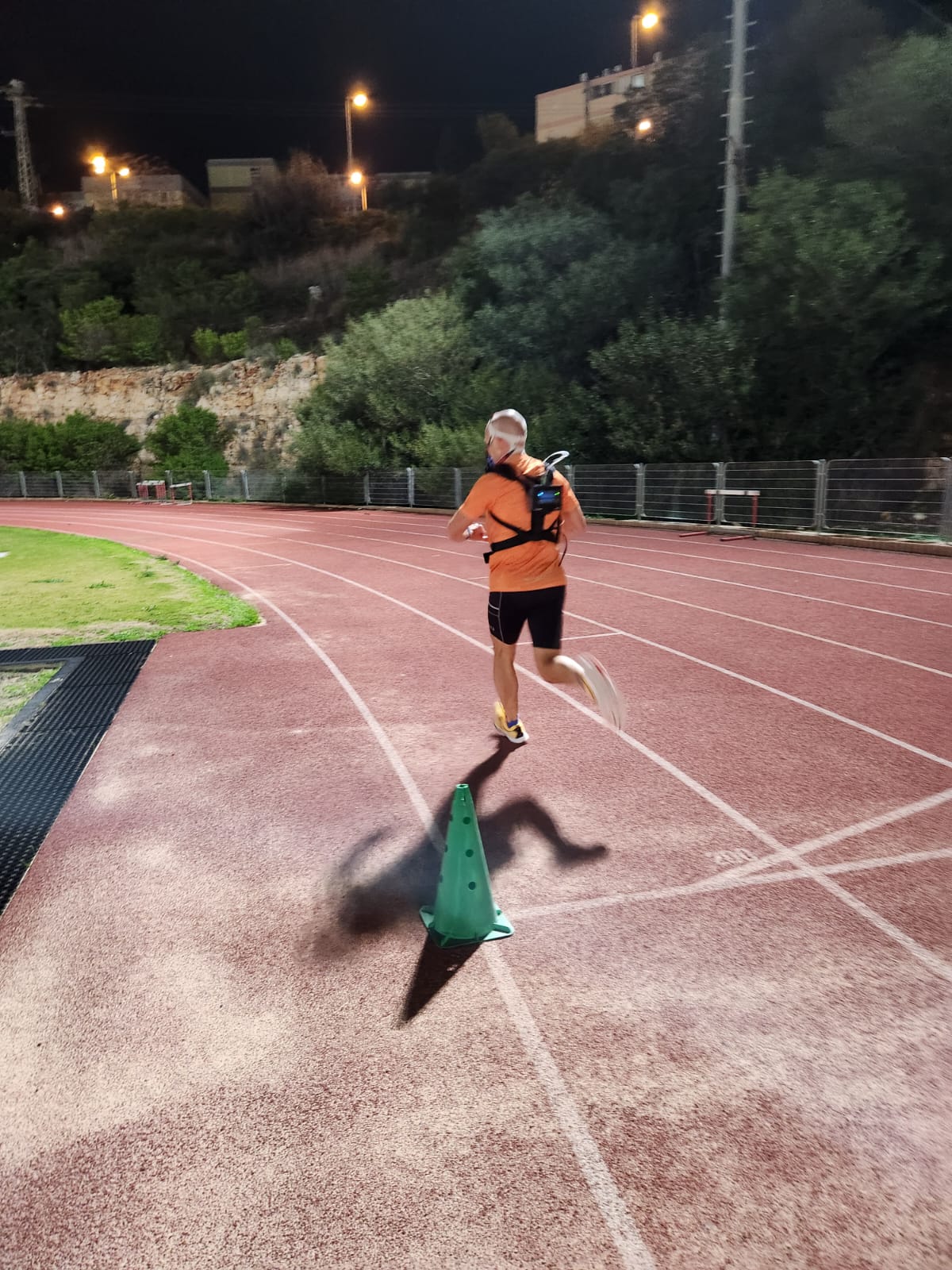}
\end{subfigure}%
\hspace{0.01\columnwidth}%
\begin{subfigure}{.24\columnwidth}
  \centering
  \captionsetup{justification=centering}
  \includegraphics[width=1\textwidth]{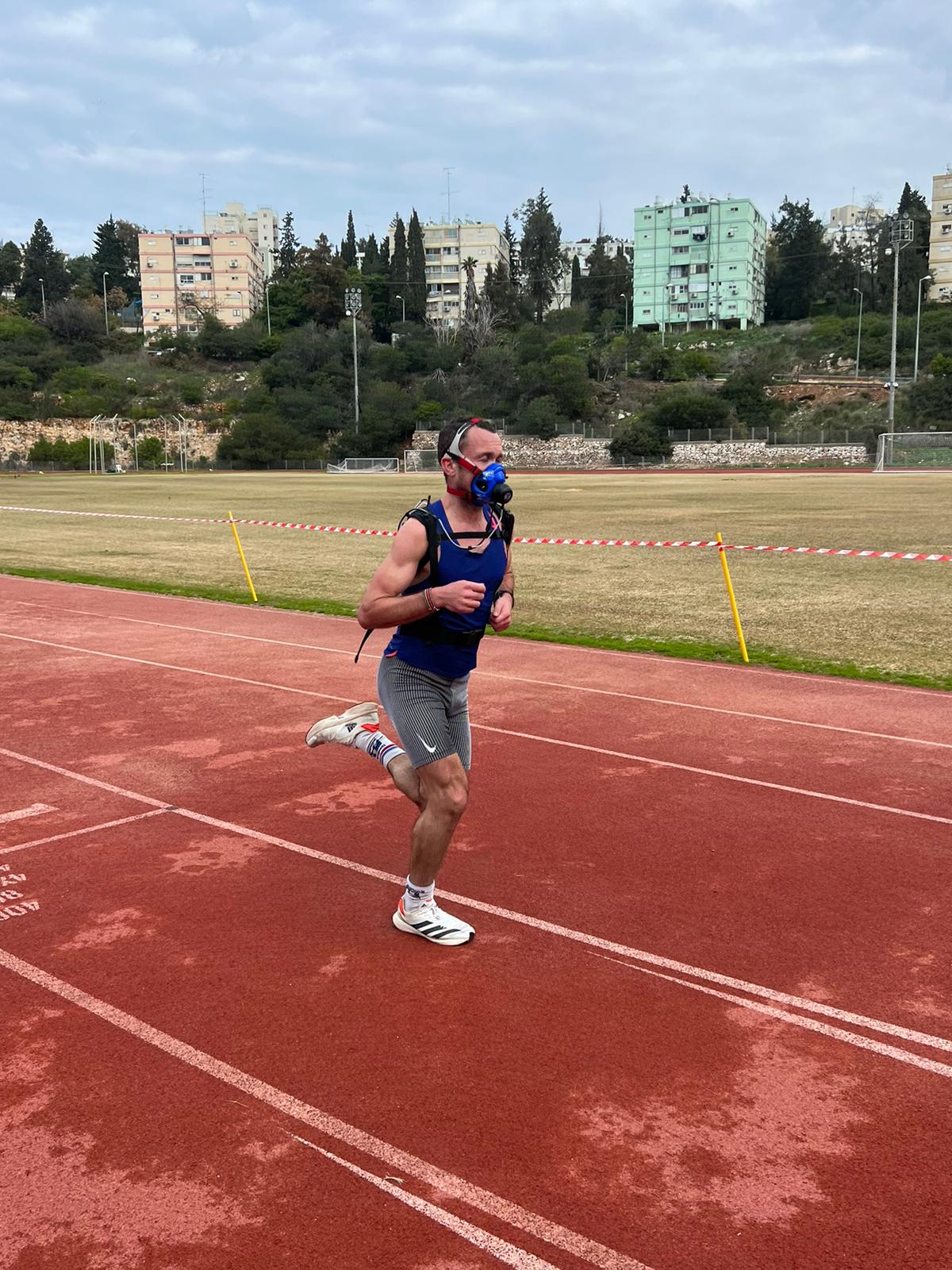}
\end{subfigure}%
\hspace{0.01\columnwidth}%
\begin{subfigure}{.24\columnwidth}
  \centering
  \captionsetup{justification=centering}
  \includegraphics[width=1\textwidth]{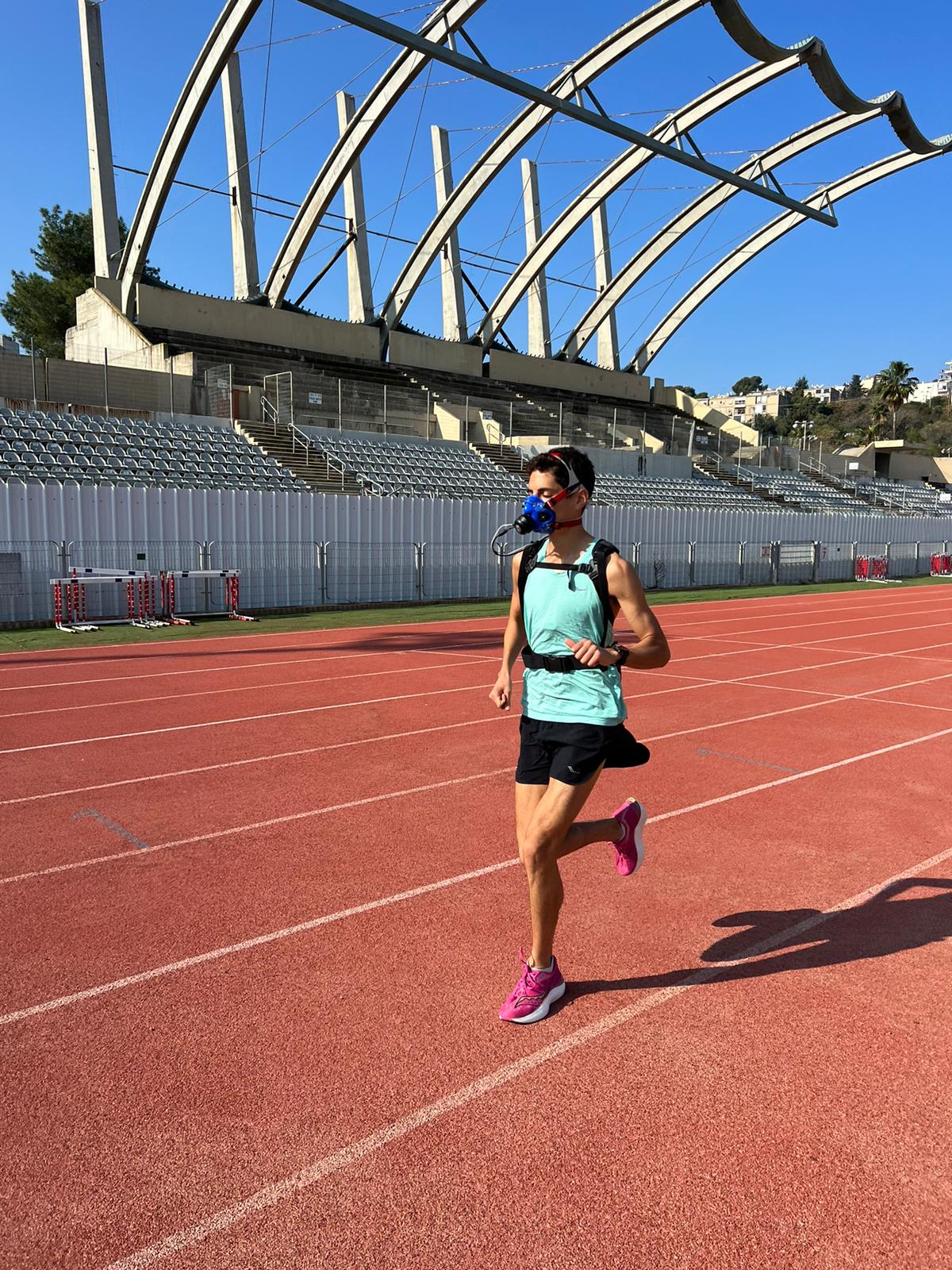}
\end{subfigure}
\Description{This figure displays four participants engaged in the experimental protocol while wearing the complete measurement apparatus. Each runner is equipped with a Garmin 965 smartwatch on their wrist, a chest-mounted HR monitor, and the Cosmed K5 portable metabolic system with its characteristic harness and facemask for respiratory gas collection. The images were captured during different experimental sessions conducted on a standard athletics track, demonstrating the real-world conditions under which the physiological data were collected. These photographs illustrate the practical implementation of our synchronized multimodal data collection methodology, which aligns laboratory-grade metabolic measurements with consumer wearable outputs.}
\label{fig:COSMED-GARMIN}
\vspace{-20pt} 
\end{figure}
Our research methodology required comprehensive physiological data capturing both laboratory-grade measurements and consumer wearable outputs. We used two complementary datasets: (1) a synchronized, multimodal dataset collected specifically for VO$_{2}$ predictions, providing aligned measurements from both laboratory and consumer devices, and (2) a more extensive historical HR dataset for training robust HR dynamics models, offering the breadth and diversity needed for developing generalizable models. Below, we describe the collection protocols and key characteristics of each dataset.

\textbf{Synchronized Multimodal Dataset:} We implemented a comprehensive protocol to gather synchronized physiological and biomechanical data from experienced runners during controlled exercise tests. Conducted at two athletic facilities, the study involved 10 participants who each completed two structured sessions. We recruited runners aged 20--50 with demonstrated high-level performance (top 1\% 10\,km race times for their age groups). This focus on highly trained athletes was intended to reduce physiological variance, facilitating more precise modeling of exercise responses. Before participation, each runner obtained medical clearance from a certified sports physician to confirm eligibility for high-intensity testing.

The protocol comprised two sessions targeting different aspects of physiological performance. In Session One (Maximal Capacity Assessment), participants performed a 1500-meter maximal-effort run to determine VO$_{2}$max. We measured metabolic data with a portable Cosmed K5 system, capturing O$_{2}$ and CO$_{2}$ exchange rates, while simultaneously recording HR and biomechanical data via a Garmin 965 smartwatch. VO$_{2}$max was calculated as the 30-second peak average VO$_{2}$, and mean running speed was computed for the entire distance. In Session Two (Incremental Testing), participants followed an individualized protocol based on Session One results, running repeated 1200-meter sets at progressively higher speeds (approximately 5\% speed increase per set) until exhaustion. To preserve physiological response patterns vital to our modeling, blood lactate samples were obtained between sets with brief (10-second) interruptions, minimizing disruptions to the overall protocol.

Data collection employed a multi-device synchronized measurement system consisting of: (1) a wrist-mounted Garmin 965 smartwatch for continuous activity monitoring; (2) a chest-mounted Garmin HR monitor capturing both HR and running dynamics (pace, cadence, vertical oscillation, altitude, stance time, vertical ratio, and step length); (3) a portable Cosmed K5 metabolic system (300g) with dedicated harness and face mask for breath-by-breath gas exchange measurement; and (4) a two-to-four-camera video system capturing biomechanical data at 200-meter intervals from dual angles throughout the athletic stadium. This multi-sensor approach ensured comprehensive data capture across both consumer-grade and laboratory-grade measurement systems, creating a synchronized dataset with precise alignment between laboratory and wearable measurements. Figure~\ref{fig:COSMED-GARMIN} illustrates participants during the protocol while wearing the integrated measurement equipment. 

\textbf{HR Modeling Dataset:} While the synchronized dataset supported VO$_{2}$ modeling, we developed our HR dynamics models using a separate, substantially more extensive dataset. This comprehensive resource comprises 831 running sessions from 20 distinct runners collected over a seven-year period (2018-2025), totaling 52,824 minutes (approximately 880 hours) and 10,222\,km of running activity, with more than 3.1 million HR data points. The scale and diversity of this dataset provided a robust foundation for capturing complex cardiovascular dynamics across various running conditions, intensities, and individual physiological profiles. Table~\ref{tab:runner_dataset_stats} in Appendix~\ref{RUNDATASET} presents detailed statistics for each participant, including session counts, accumulated training time, and total distance covered.

We deliberately chose chest strap HR monitors over wrist-worn devices for our HR data collection due to their superior measurement accuracy during exercise. Unlike wrist-based smartwatches that use photoplethysmography (PPG)—an optical method measuring blood flow through optical sensors—chest straps capture electrical signals directly from the heart via electrocardiogram (ECG), providing precise beat-to-beat measurements even during intense physical activity. The limitations of wrist-worn PPG sensors are well-documented: movement artifacts, sweat interference, and reduced peripheral blood flow during high-intensity exercise can compromise optical sensors' signal integrity~\cite{sartor2018wrist,parak2021comparison,martin2023activity}. In contrast, ECG-based chest straps maintain consistent accuracy across varying exercise intensities, making them the preferred choice for physiological research requiring high temporal resolution and reliability.\label{dataset}
\section{Modeling Heart Rate Dynamics}\label{Framework1}
Accurately modeling HR dynamics during physical activity represents a foundational step in understanding physiological responses to exercise. Beyond its inherent value, precise HR estimation addresses three critical challenges in exercise physiology and wearable technology: (1) it enables exploration of how biomechanical features directly influence physiological responses, potentially revealing new relationships between movement patterns and cardiovascular demand; (2) it provides robust solutions for signal interruption and interpolation—persistent problems in wearable devices during high-intensity activities; and (3) it leverages established physiological correlations between HR, power output, and VO$_{2}$ to enable more sophisticated training insights. 
 
In this section, we introduce two complementary approaches for HR prediction during running, each offering unique advantages: a physiologically-constrained ordinary differential equation (ODE) approach that mathematically models cardiac adaptation rates, and a Kalman filtering framework that optimally balances prior physiological estimates with new observations.
\begin{figure*}[t!]
\centering
\caption{Schematic of HR prediction framework. Wearable data is encoded into latent states $\mathbf{s}_t$ and processed via either: (1) neural ODE solver with equation $\frac{dg_t}{dt} = d_t-g_t$, or (2) Kalman filter with learnable updates. Both use estimated moments $\hat{\mu}_t$ and $\hat{\sigma}_t$ for denormalization to produce HR predictions $\hat{h}_t$. Gray indicates learnable parameters $\bm{\mathrm{\vartheta}}_\ast$.}
\Description{The figure illustrates the complete computational pipeline of the HR prediction framework. On the left, multimodal wearable measurements (shown as waveforms in different colors) serve as inputs, passing through a normalization block, then an encoder to produce encoded features, followed by a temporal model that outputs latent state vectors. The pipeline then branches into two alternative approaches: (1) An ODE solver, and (2) A Kalman filter architecture (bottom branch) with its mathematical formulation including state transitions, Kalman gain, and covariance updates. Both approaches converge at a denormalization step on the right side that produces the final HR predictions. All learnable parameters are highlighted in gray throughout the diagram, showing where the model can be trained.}
\includegraphics[width=0.85\linewidth]{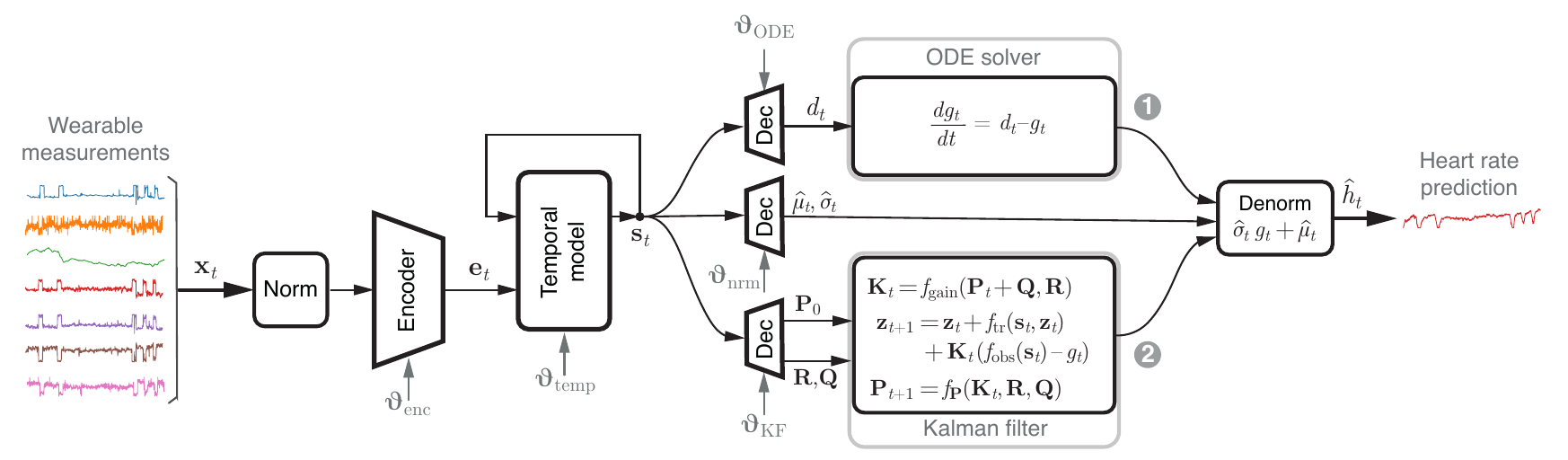}
\label{fig:model_architecture}
\end{figure*}

\textbf{Problem Formulation.} Our goal is to predict HR during running, given a multivariate time series of biomechanical and environmental features. Let $X = \{x_1, x_2, \ldots, x_T\}$ be the sequence of external parameters (e.g., pace, cadence, vertical oscillation, altitude, stance time, vertical ratio, step length), where each $x_t \in \mathbb{R}^d$ describes running dynamics at second $t$. We aim to produce the corresponding HR sequence $Y = \{y_1, y_2, \ldots, y_T\}$ at 1-second intervals. The main challenge is capturing both rapid cardiac responses to intensity changes and slower physiological adaptations, all within biologically plausible limits. 

\textbf{Data Preprocessing.} We used the HR modeling dataset (Section~\ref{dataset}), comprising over 800 running sessions from 20 runners (2018--2025). Raw FIT files were segmented into 60-second, non-overlapping windows, with domain-specific transformations applied to maintain physiological relevance. Pace (m/s) was converted to sec/km, vertical oscillation was normalized by each runner's height, and stance time (percent) was converted to a fraction. Step length was scaled from millimeters, altitude data was split into absolute values and relative gains to capture cardiovascular demand influences, and cadence was doubled to reflect full running cycles. These transformations preserved physiological interpretability and ensured proper input scaling.

\textbf{Model Architecture.} Our framework combines neural feature extraction with domain-specific models of cardiac dynamics as described in Figure~\ref{fig:model_architecture}. We denote learnable functions by $f$ and their parameters by $\bm{\mathrm{\vartheta}}$, with subscripts indicating the specific part of the system.

The model receives a workout window $\mathbf{x}_t$ spanning the interval $[t-T,t]$. The window is encoded into a latent space by an encoder, $\mathbf{e}_t = f_\mathrm{enc}(\mathbf{x}_t; \bm{\mathrm{\vartheta}}_\mathrm{enc})$, followed by an auto-regressive model $\mathbf{s}_t = f_\mathrm{temp}(\mathbf{e}_t, \mathbf{s}_{t-T} ; \bm{\mathrm{\vartheta}}_\mathrm{temp})$ with $\mathbf{s}_0 = \mathbf{0}$. Two dynamic models predict the latent timeseries $g_t$, from which the predicted HR is decoded using 
$
\hat{h}_t = f_{\mathrm{dec}} (g_t, \mathbf{s}_t ; \bm{\mathrm{\vartheta}}_\mathrm{dec}).
$
This backbone is implemented with a fully connected feature encoder with leaky ReLU activation for $f_\mathrm{enc}$, and a gated recurrent unit (GRU) for $f_\mathrm{temp}$. For decoding HR from $g_t$, we use an affine mapping $\hat{h}_t = \hat{\sigma}_t g_t + \hat{\mu}_t$, where $\mu_t$ and $\sigma_t$ are the HR mean and standard deviation in the interval $[t-T,t]$ estimated from the latent state by fully-connected models $\hat{\mu}_t = \mu(\mathbf{s}_t; \bm{\mathrm{\vartheta}}_\mathrm{nrm})$ and $\hat{\sigma}_t = \sigma_t(\mathbf{s}_t; \bm{\mathrm{\vartheta}}_\mathrm{nrm})$. For additional details, refer to Appendix~\ref{HRModelAppendix}.

We now describe our two alternative approaches for dynamic HR modeling:
\textbf{ODE-based dynamic model.} Our first model for the latent dynamics is based on a non-linear first-order continuous-time ODE,
$
\label{eq:hr_dynamics}
\frac{\partial{g_t}}{\partial{t}} = f_{\text{ODE}}(g_t, \mathbf{s}_t; \bm{\mathrm{\vartheta}}_\mathrm{ODE}),
$
which we integrate with a fourth-order Runge-Kutta (RK4) method.
As the right-hand side of the ODE, we used $f_{\text{ODE}} = d_t - g_t$, where $d_t = d(\mathbf{s}_t; \bm{\mathrm{\vartheta}}_\mathrm{dec})$ is a fully-connected network estimating the demand from the latent state $\mathbf{s}_t$. 

\textbf{Kalman Filter Approach.} Our second model for the latent dynamics is a neural version of an enhanced Kalman filter~\citep{kalman1960new}. 
We define a two-dimensional state vector $\mathbf{z}_t = (g_t, \dot{g}_t)$ comprising the latent HR and its velocity.
A scalar Kalman gain is calculated using a neural network 
$
k_t = f_\mathrm{gain}(\mathbf{P}_{t}+\mathbf{Q}, \mathbf{R})
$
and is used for the \emph{a posteriori} state update
$
\mathbf{z}_{t+1} = \mathbf{z}_{t} + 
f_{\text{tr}}(\mathbf{s}_t, \mathbf{z}_{t}; \bm{\mathrm{\vartheta}}_{\text{tr}}) + k_t \left(
f_{\text{obs}}(\mathbf{s}_t; \bm{\mathrm{\vartheta}}_{\text{obs}}) - g_t
\right) (1, \gamma_t), \nonumber
$
mimicking the standard Kalman filter.

The scaling factor $\gamma_t$ for velocity updates is set to 0.5, implementing a differential relationship between HR and velocity corrections. This reduced influence on velocity updates allows the model to maintain smoother trajectory changes while still responding to new observations, effectively damping oscillations that might occur from measurement noise.

The covariance update $\mathbf{P}_{t+1}$ follows a modified Kalman formulation:
$
\mathbf{P}_{t+1} = \left(\mathbf{P}_{t} + \mathbf{Q} \right) \odot
\begin{pmatrix} 
1 - k_t & 0 \\
0 & 1 - \gamma_t k_t  \nonumber
\end{pmatrix},
$
where the diagonal structure preserves computational efficiency while allowing independent uncertainty tracking for both HR value and velocity components.

The decoder neural networks provide the complete set of Kalman filter parameters: initial state $\mathbf{z}_0 = (g_0, \dot{g}_0)$, initial covariance matrix $\mathbf{P}_0 = \text{diag}(\sigma^2_{g_0}, \sigma^2_{\dot{g}_0})$, process noise covariance $\mathbf{Q} = \text{diag}(\sigma^2_{\text{proc},g}, \sigma^2_{\text{proc},\dot{g}})$, and measurement noise variance $\mathbf{R} = \sigma^2_{\text{meas}}$. These parameters are computed from the final hidden state of the GRU encoder, allowing the model to adapt its filtering behavior to different individuals and physiological conditions.

We used fully-connected models for $f_{\text{enc}}$, $f_{\text{tr}}$, $f_{\text{obs}}$, and $f_{\text{gain}}$. The gain function captures prediction errors made by the observation function, assigning greater weight to new measurements when errors are large. Additional details are provided in Appendix~\ref{KalmanAppendix}.

\textbf{Training Methodology.} Both models are trained fully-supervised to minimize the Mean Absolute Error (MAE) between predicted and true HR. To ensure physiological realism, we added coarse-scale regularization terms supervising the HR first- and second-order statistics, 
$
\mathcal{L} = \mathbb{E}_t | \hat{h}_t - h_t | + \lambda \mathbb{E}_t |\hat{\mu}_t - \mu_t | + \lambda \mathbb{E}_t | \hat{\sigma}_t - \sigma_t |,
$
Here, $\mathbb{E}_t$ denotes temporal expectation (in practice, finite-sample average on the training set), $h_t$ is the ground-truth HR, and $\mu_t = \int_{t-T}^t g_\tau d\tau$ and $\sigma_t^2 = \int_{t-T}^t (g_\tau-\mu_\tau)^2 d\tau$ are the ground-truth first- and second-order moments used for the supervision. The parameter $\lambda$ controls the influence of regularization by temporal statistics and was set to $\lambda=0.1$ following an ablation study. For the ODE model, we backpropagate through the neural ODE solver using the adjoint method as proposed~\cite{chen2018neural}.

We tested generalizability via runner-based leave-three-out cross-validation, ensuring no overlap between training and evaluation participants. Models were assessed in eight splits, each with more than 110 sessions (20--140 minutes per session). We compared two network architectures—128 neurons (2 layers) vs.\ 64 neurons (3 layers)—under consistent settings (batch size = 64). An adaptive learning rate scheduler and early stopping reduced overfitting. Appendix~\ref{Appex:TrainingHR} discusses additional information.
\begin{table}[t]
\centering
\footnotesize
\caption{Performance comparison of HR prediction models using Standard/Generative approaches. Values shown as "Standard/Generative" format for the four models.}
\label{tab:results_performance}
\begin{tabular}{lcccc}
\hline
\textbf{Metric} & \textbf{Kalman} & \textbf{Kalman} & \textbf{ODE} & \textbf{ODE} \\
 & \textbf{(128,2)} & \textbf{(64,3)} & \textbf{(128,2)} & \textbf{(64,3)} \\
\hline
\multicolumn{5}{l}{\textbf{Overall Performance}} \\
\hline
MAE (bpm) & \textbf{2.81}/11.70 & 3.01/11.12 & 2.84/12.79 & 2.91/12.52 \\
RMSE (bpm) & \textbf{4.60}/13.98 & 4.96/13.45 & 4.70/15.22 & 4.87/14.75 \\
MAPE (\%) & \textbf{2.17}/8.49 & 2.39/8.35 & 2.18/9.18 & 2.24/8.84 \\
Correlation & \textbf{0.87}/0.46 & 0.86/0.45 & 0.87/0.47 & 0.86/0.47 \\
R$^2$ & \textbf{0.73}/-2.51 & 0.71/-1.59 & 0.72/-2.41 & 0.70/-1.95 \\
Mean Diff. (bpm) & 0.02/-2.41 & 0.05/-1.69 & 0.01/-2.32 & \textbf{-0.02}/-6.60 \\
StdDev Diff. (bpm) & \textbf{4.55}/10.13 & 4.92/10.42 & 4.67/10.78 & 4.83/9.81 \\
\hline
\multicolumn{5}{l}{\textbf{Performance by HR Zone (MAE in bpm)}} \\
\hline
Low HR & \textbf{10.78}/15.12 & 13.24/18.82 & 10.91/14.65 & 12.22/13.92 \\
Medium HR & \textbf{3.05}/9.82 & 3.22/9.65 & 3.13/10.74 & 3.22/9.69 \\
High HR & \textbf{1.99}/12.22 & 2.09/11.87 & 2.02/13.24 & 2.09/15.20 \\
\hline
\multicolumn{5}{l}{\textbf{Performance by HR Stability (MAE in bpm)}} \\
\hline
Transitions & \textbf{10.50}/15.05 & 11.93/16.52 & 11.44/16.36 & 11.64/14.75 \\
Steady-State & \textbf{2.78}/11.68 & 2.99/11.11 & 2.81/12.78 & 2.89/12.51 \\
\hline
Avg. sessions/split & 125/125 & 113/113 & 122/122 & 122/122 \\
\hline 
\end{tabular} 
\end{table}

\textbf{Results.} Table~\ref{tab:results_performance} presents a comprehensive evaluation of our HR prediction models across different architectures and inference settings. The table compares Standard inference (where the first second of each window is known) versus Generative Sequencing (where only the first second of the entire session is provided) across Kalman and ODE models with varying architectures. Our approach offers advantages over prior work: Apple's approach~\cite{nazaret2023modeling} used 10-second intervals with proprietary data, whereas we provide fine-grained 1-second predictions. Our open-source implementations enhance reproducibility and enable direct comparisons. Crucially, \textbf{our approach requires no historical data} from previous sessions, making it suitable for new runners without a calibration phase. 

\textbf{Under standard inference}, the Kalman (128,2) model achieved the best overall performance (MAE: 2.81\,bpm, RMSE: 4.60\,bpm, MAPE: 2.17\%, correlation: 0.87, R$^2$=0.73). The ODE (128,2) model followed closely at 2.84\,bpm MAE. This accuracy is particularly notable given the diversity of running intensities in our dataset. \textbf{Under Generative Sequencing}, using only the first second of the first window to drive predictions across an entire session—all models demonstrated consistent performance over extended durations, though with an expected decrease without periodic recalibration. Analysis by HR zone and stability state revealed strong results in steady-state and high-intensity scenarios, with MAE as low as 1.99\,bpm in high HR zones. This outcome has significant practical importance for wearable devices prone to signal disruptions during exercise.
\begin{figure}[t!]
\centering
\vspace{-12pt} 
\caption{Example HR prediction from the ODE based model (128, 2) for a high intensity workout. Blue line represents true HR measurements, and red line shows model predictions.}
\includegraphics[width=1\linewidth]{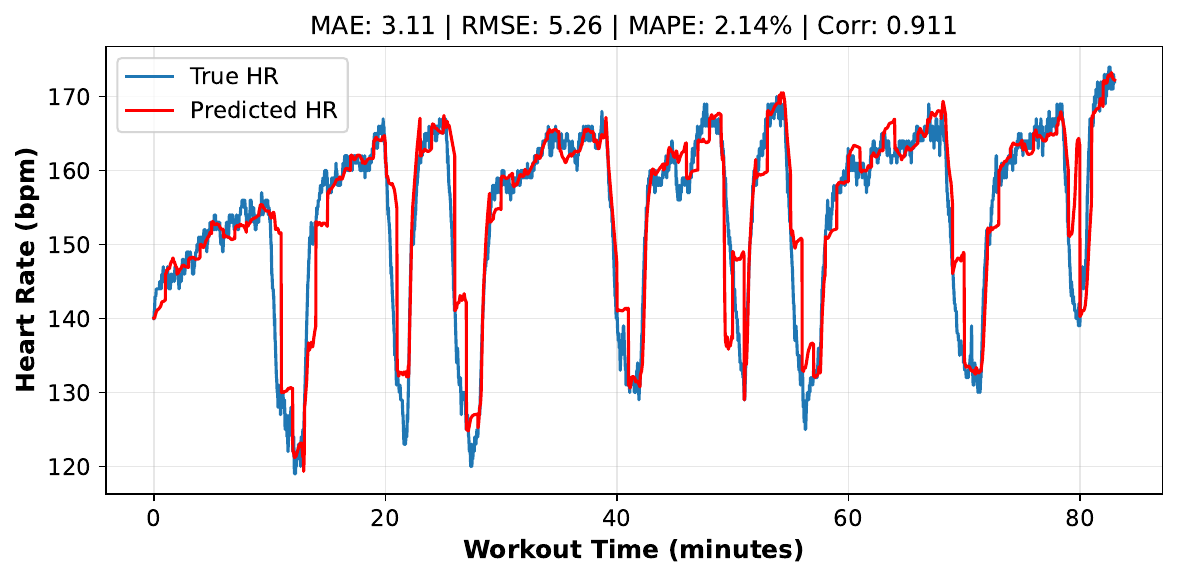}
\Description{This figure shows a time series plot comparing actual HR measurements (blue line) with model predictions (red line) during a high-intensity running session.}
\label{fig:results_plots}
\vspace{-12pt} 
\end{figure}
Figure~\ref{fig:results_plots} illustrates ODE-based model (128,2) performance for a high-intensity interval session (HR: 120-170\,bpm). The model achieves exceptional accuracy (MAE: 1.68\,bpm, correlation: 0.933) despite challenging physiological transitions, effectively capturing both general trends and fine-grained HR dynamics. Additional visualizations of different workout intensities and prediction modes appear in Appendix~\ref{HRADDITIONALFIGURES}.

These visualizations confirm our models effectively capture both general trends and fine-grained HR dynamics across diverse running conditions. Intermittent recalibration (e.g., at window boundaries) could harness Standard inference accuracy while addressing real-world challenges like PPG signal loss in consumer wearables.

\section{Predicting Instantaneous Oxygen Consumption from Consumer Wearables}\label{Framework2}
Accurately estimating VO$_{2}$ during exercise is essential for understanding energy expenditure, quantifying training adaptation, and assessing overall fitness~\citep{american2013acsm}. However, conventional VO$_{2}$ measurement methods rely on specialized metabolic carts and face masks for respiratory gas collection—equipment that typically costs over 30,000\$ and requires expert operation. These technical and financial constraints fundamentally restrict practical use in real-world, continuous-monitoring scenarios where most exercise actually occurs.

To overcome this critical methodological limitation, we introduce what we believe to be the first method for predicting instantaneous VO$_{2}$ trajectories solely from consumer-grade wearable sensor data. Our approach eliminates the need for costly respiratory gas analysis while maintaining clinically acceptable accuracy, effectively democratizing access to sophisticated metabolic insights previously confined to laboratory settings.

\textbf{Problem Formulation.} We aim to predict VO$_{2}$ during running using a multivariate time series of biomechanical and environmental features. Let $X = \{x_1, x_2, \ldots, x_T\}$ represent a sequence of external parameters including pace, cadence, vertical oscillation, altitude, stance time, vertical ratio, and step length, where each $x_t \in \mathbb{R}^d$ captures running dynamics at second $t$. Our objective is to generate a corresponding VO$_{2}$ sequence $Y = \{y_1, y_2, \ldots, y_T\}$ at 1-second resolution. The primary challenge lies in accurately modeling both rapid metabolic responses to intensity changes and slower physiological adaptations while maintaining biologically plausible constraints.

\textbf{Data Preprocessing.} Our methodology leverages a precisely synchronized multimodal dataset (Section~\ref{dataset}) establishing direct relationships between laboratory-grade metabolic measurements and wearable device outputs. We paired breath-by-breath data from a Cosmed K5 metabolic analyzer—the gold standard for VO$_{2}$ measurement~\citep{perez2018accuracy}—with physiological and biomechanical metrics from a Garmin 965 smartwatch and chest-mounted HR monitor. Temporal alignment was achieved through a zero-order hold on Cosmed signals and session synchronization via GPS coordinates and event markers. 

Sessions were segmented into 60-second windows coupling Cosmed and smartwatch data, creating a comprehensive dataset where metabolic parameters (VO$_{2}$, VCO$_{2}$, ventilation, RER) sampled at 0.3--0.5\,Hz precisely align with smartwatch features (HR, cadence, vertical oscillation, altitude) at 1\,Hz. To mitigate breath-by-breath variability, we applied a Savitzky-Golay filter (15-sample window, polynomial order 3)~\citep{savitzky1964smoothing} that preserves rapid physiological transitions while removing measurement artifacts.

From the refined dataset, we derived a comprehensive feature set capturing the multifaceted nature of exercise physiology. The features include cardiac measurements like HR, biomechanical indicators such as cadence, vertical oscillation ratio, step length, and stance time, contextual variables including pace, grade, and cumulative distance, positional encodings of session window index and total elapsed time, and anthropometric data like age, gender, height, and weight. This diverse feature set enables our model to account for immediate cardiorespiratory responses to exercise intensity transitions and individual physiological characteristics, establishing a robust foundation for accurate VO$_{2}$ estimation using only consumer-grade wearable data.
\begin{figure*}[t!]
\centering
\caption{The VO$_{2}$ prediction framework. The pipeline processes wearable sensor inputs through normalization and encoding steps before implementing a dual-stream architecture with a neural Kalman filter and direct estimation pathway, which are adaptively blended to produce the final VO$_{2}$ predictions.}
\Description{The figure illustrates the VO2 prediction framework as a left-to-right flowchart. On the left, multimodal wearable sensor data (shown as colored waveforms) enter the system and pass through a normalization block. Data then flows through an encoder and temporal model, creating latent representations. The architecture then splits into three parallel pathways: (1) a direct prediction branch at the top, (2) a Kalman filter implementation in the middle showing mathematical equations for state updates, adaptive gains, and covariance calculations, and (3) a decoder pathway at the bottom. The outputs from the direct and Kalman branches are combined through an adaptive blending module, and finally pass through a denormalization step to produce the VO2 estimates. Learnable parameters are indicated throughout the diagram, showing where the model can adapt to individual physiological differences.}
\includegraphics[width=0.8\linewidth]{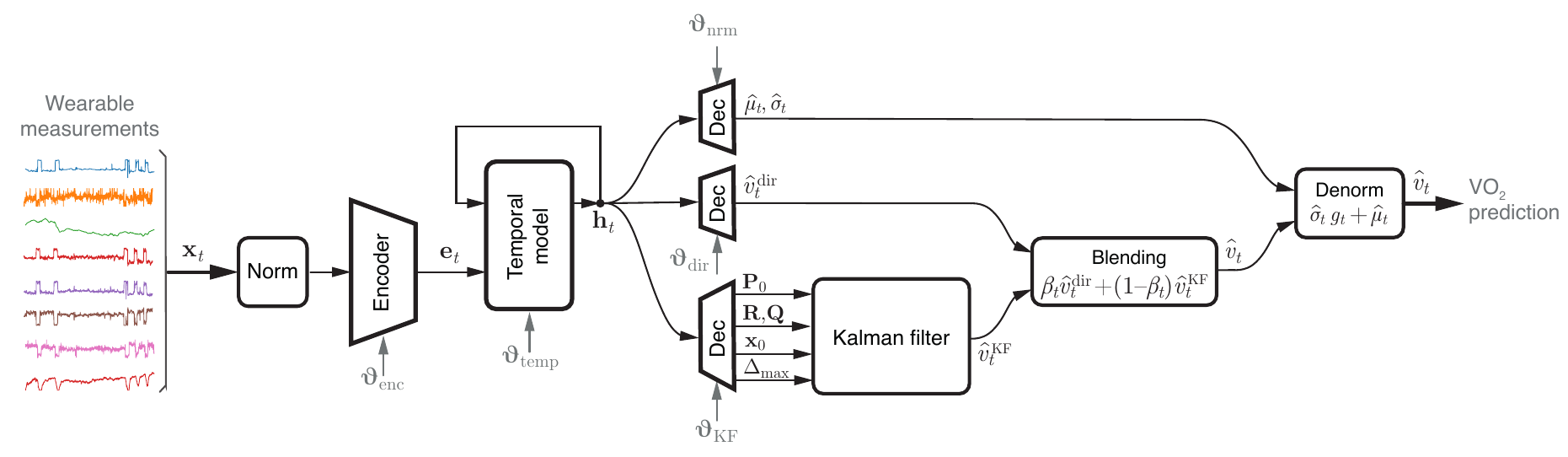}
\label{fig:VO2model_architecture}
\end{figure*}

\textbf{Model Architecture.} Predicting VO$_{2}$ during exercise presents unique challenges due to the complex interplay between physiological systems and rapid metabolic adaptations. We model VO$_{2}$ dynamics as a state–estimation problem subject to physiologically informed constraints, denoting the latent oxygen‑consumption state at time~$t$ by $v_t$ (and $\hat{v}_t,\tilde{v}_{t^-}$ for its predicted and trend‑extrapolated versions, respectively). Figure~\ref{fig:VO2model_architecture} summarizes the framework. In what follows, learnable functions are written $f(\,\cdot\,;\bm{\vartheta})$, where the subscript on~$\bm{\vartheta}$ indicates the module.

The model processes a workout window $\mathbf{x}_{t}$ spanning $[t-T,t]$ containing ten biomechanical, HR, and temporal features. A feature encoder and bidirectional GRU generate a hidden representation
$
\mathbf{h}_{t}=f_{\text{temp}}\!\bigl(f_{\text{enc}}(\mathbf{x}_{t};\bm{\vartheta}_{\text{enc}});\bm{\vartheta}_{\text{temp}}\bigr).
$

\paragraph{Neural–Kalman update.}
Using $\mathbf{h}_{t}$, the neural Kalman filter generates a Kalman filter-based VO$_2$ estimate $\hat{v}_{t}^{\text{KF}}$ through an adaptive mechanism that balances new observations with prior estimates:
\begin{equation}
\hat{v}_{t}^{\text{KF}} = \hat{v}_{t-1}
+ f_{\text{gain}}(\mathbf{h}_{t};\bm{\vartheta}_{\text{KF}})
\bigl[\,f_{\text{obs}}(\mathbf{h}_{t};\bm{\vartheta}_{\text{obs}})
      - \hat{v}_{t-1}\bigr]_{%
      f_{\Delta_{\min}}(\mathbf{h}_{t};\bm{\vartheta}_{\Delta})}^{%
      f_{\Delta_{\max}}(\mathbf{h}_{t};\bm{\vartheta}_{\Delta})}, \nonumber
\end{equation}
where $f_{\text{gain}}$ modulates observation trust and $f_{\Delta_{\min}},f_{\Delta_{\max}}$ enforce physiologically constrained rate-of-change limits.

\paragraph{Trend/direct dual pathway.}
To enhance robustness, we employ a dual-pathway approach blending trend-extrapolated and direct neural estimates:
\begin{align}
\tilde{v}_{t^-} &=
     \hat{v}_{t-1}
     + f_{\text{trend}}(\mathbf{h}_{t};\bm{\vartheta}_{\text{trend}}) \nonumber
       \bigl(\hat{v}_{t-1}-\hat{v}_{t-2}\bigr), \\ \nonumber
\hat{v}_{t}^{\text{dir}} &= f_{\text{direct}}(\mathbf{h}_{t};\bm{\vartheta}_{\text{dir}}), \\ \nonumber
\hat{v}_{t} &= \beta_{t} \hat{v}_{t}^{\text{dir}} + (1-\beta_{t}) \hat{v}_{t}^{\text{KF}},
\end{align}
where $\beta_{t} = f_{\text{blend}}(\mathbf{h}_{t};\bm{\vartheta}_{\text{blend}})$ determines the blending proportions. All modules $f_{\text{enc}}$, $f_{\text{gain}}$, $f_{\text{obs}}$, $f_{\Delta_{\min}}$, $f_{\Delta_{\max}}$, $f_{\text{trend}}$, $f_{\text{blend}}$, and $f_{\text{direct}}$ are implemented as multilayer perceptrons (MLPs) with corresponding parameters $\bm{\vartheta}_{\text{enc}}$, $\bm{\vartheta}_{\text{KF}}$, $\bm{\vartheta}_{\text{obs}}$, $\bm{\vartheta}_{\Delta}$, $\bm{\vartheta}_{\text{trend}}$, $\bm{\vartheta}_{\text{blend}}$, and $\bm{\vartheta}_{\text{dir}}$, respectively. The temporal model $f_{\text{temp}}$ is a bidirectional GRU (biGRU) with parameters $\bm{\vartheta}_{\text{temp}}$ that captures temporal dynamics. The final output is normalized using parameters $\bm{\vartheta}_{\text{nrm}}$. Detailed layer configurations are provided in Appendix~\ref{KalmanVO2Appendix}.

\textbf{Training Methodology.} Our model is trained fully-supervised to minimize a composite loss function that balances immediate prediction accuracy with physiological plausibility,
\begin{align}
\mathcal{L} &= \mathbb{E}_t | \hat{v}_{t} - v_t | + \lambda_{\text{dynamic}} \mathbb{E}_t \Big( \alpha \Big| \frac{d\hat{v}_{t}}{dt} - \frac{dv_t}{dt} \Big| + \beta \Big| \nonumber\frac{d^2\hat{v}_{t}}{dt^2} - \frac{d^2v_t}{dt^2} \Big| \Big) \\ \nonumber
&+ \lambda_{\text{aux}} \Big( \mathbb{E}_t | \hat{\mu}_t - \mu_t | + \mathbb{E}_t | \hat{\sigma}_t - \sigma_t | + \mathbb{E}_t | \hat{\Delta}_t - \Delta_t | \Big),\nonumber
\end{align}
where $\hat{v}_{t}$ represents the blended estimate $\beta_{t} \hat{v}_{t}^{\text{dir}} + (1-\beta_{t}) \hat{v}_{t}^{\text{KF}}$. Here $\mathbb{E}_t$ denotes temporal expectation, $v_t$ is the ground-truth VO$_2$ value, $\frac{dv_t}{dt}$ and $\frac{d^2v_t}{dt^2}$ represent the first and second derivatives of the VO$_2$ signal capturing velocity and acceleration dynamics. The coefficients $\alpha$ and $\beta$ weight the relative importance of the derivatives. For the statistical terms, $\mu_t$ is the temporal mean, $\sigma_t$ is the standard deviation, and $\Delta_t = Q_{0.95}(|v_\tau - v_{\tau-1}|)$ represents the 95th percentile of absolute differences in consecutive VO$_2$ values. The parameters $\lambda_{\text{dynamic}}$ and $\lambda_{\text{aux}}$ follow a curriculum-based schedule, with $\lambda_{\text{dynamic}}$ increasing from 0 to 0.7 and $\lambda_{\text{aux}}$ increasing from 0.1 to 0.3 over the course of training to gradually emphasize physiological realism.

We complement this curriculum learning with adaptive gradient clipping that permits larger parameter updates during early training stages while enforcing stability as the model converges. This progressive optimization strategy enables the model to establish foundational prediction capabilities before refining its representation of the nuanced transition dynamics that characterize VO$_{2}$ response during variable-intensity exercise. Appendix~\ref{trainingVO2Appendix} contains additional information.

\textbf{Results.} We evaluated the model's generalizability using a leave-one-runner-out approach, training on data from all but one participant and testing on the remaining runner. Due to the dataset's limited size, we used overlapping windows during training but non-overlapping windows for left-out runner evaluation, maintaining session integrity. A unique aspect of our method is generating complete sequence predictions using only \textbf{the first second of VO$_{2}$ data} for initialization—mimicking real-world scenarios where continuous metabolic measurement is impractical. This minimal conditioning approach, potentially derived from domain knowledge or brief calibration, removes the need for ongoing respiratory gas analysis. Our sequence-to-sequence VO$_{2}$ predictions were assessed using multiple metrics, detailed in Table~\ref{tab:vo2_comparison_highlight}.

\begin{table}[!t]
\footnotesize
\centering
\caption{Seq-to-Seq VO$_{2}$ Prediction for two models using Leave-One-Runner-Out Cross-Validation, with real HR.}
\label{tab:vo2_comparison_highlight}
\begin{tabularx}{\columnwidth}{l *{6}{>{\centering\arraybackslash}X}}
\toprule
& \multicolumn{3}{c}{Model 256-2} & \multicolumn{3}{c}{Model 128-4} \\
\cmidrule(r){2-4} \cmidrule(l){5-7}
Runner & MAE & RMSE & MAPE & MAE & RMSE & MAPE \\
& & & (\%) & & & (\%) \\
\midrule
Runner-1 & 520.20 & 582.88 & 17.17 & \textbf{325.40} & \textbf{399.68} & \textbf{11.59} \\
Runner-2 & 1326.20 & 1361.69 & 35.40 & \textbf{453.74} & \textbf{493.50} & \textbf{12.81} \\
Runner-3 & 218.41 & 473.09 & 11.47 & 212.02 & 477.44 & 11.59 \\
Runner-4 & 127.55 & 198.96 & 7.85 & 129.08 & 207.55 & 7.95 \\
Runner-5 & 606.88 & 636.92 & 22.08 & 451.26 & 488.13 & \textbf{16.42} \\
Runner-6 & 261.49 & 328.66 & 13.26 & 190.94 & 270.33 & \textbf{10.02} \\
Runner-7 & 159.01 & 201.17 & 6.19 & 131.81 & 192.59 & 5.35 \\
Runner-8 & 321.80 & 382.35 & 13.54 & 235.29 & \textbf{298.0} & \textbf{10.45} \\
Runner-9 & 321.01 & 398.18 & 25.05 & 267.14 & 344.00 & \textbf{21.26} \\
Runner-10 & 108.75 & 155.70 & 6.24 & 112.37 & 160.21 & 6.54 \\
\midrule
Aggregate & 397.13 & 471.96 & 15.83 & \textbf{251.00} & \textbf{333.20} & \textbf{11.40} \\
\bottomrule
\end{tabularx}
\label{resultsVO2}
\end{table}

\textbf{Contextualizing Prediction Errors.} Our model's performance aligns with established benchmarks for instantaneous VO$_{2}$ estimation. At high intensities (VO$_{2}$ \(\sim\)3000 ml/min), gold-standard metabolic equipment typically accepts errors of 300–600 ml/min (10–20\%). Our consumer-grade smartwatch method achieves an aggregate Mean Absolute Percentage Error (MAPE) of 11.40\%, compared to the 256-2 model's 15.83\%, demonstrating remarkable accuracy given the limited input data. This error range corresponds to approximately 4.3–8.6 ml·kg\(^{-1}\)·min\(^{-1}\) for a 70 kg athlete and aligns with discrepancies observed between research-grade devices.
\begin{figure}[t]
\centering
\caption{VO$_{2}$ predictions from first-second data during an incremental testing session with periodic lactate measurements. Blue: ground truth; orange: model predictions.}
\includegraphics[width=1\linewidth]{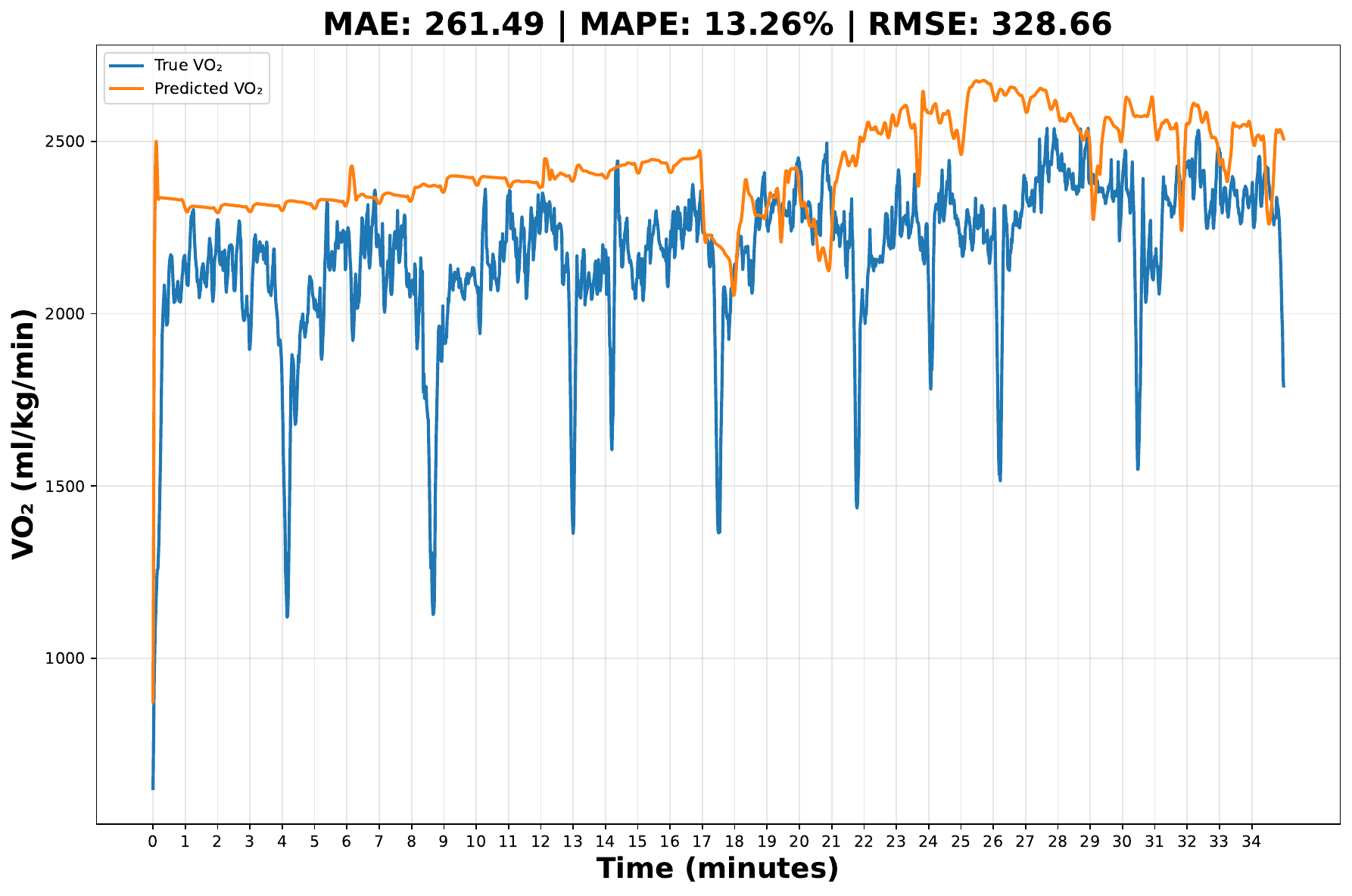}
\Description{A graph comparing predicted VO2 (orange) with ground truth measurements (blue) during a 35-minute incremental test with progressively increasing intensity and periodic drops during lactate sampling. The model accurately tracks metabolic demand using only the first second of VO2 data plus smartwatch signals.}
\label{fig:vo2_predictions} 
\vspace{-12pt}
\end{figure}

Several physiological factors inherently complicate VO$_{2}$ prediction: \textbf{breath-by-breath variability}~\citep{mcnulty2019repeat} introduces inherent ($\pm$5-10\%) fluctuations, \textbf{nonlinear VO$_{2}$ kinetics}~\citep{zoladz1995non} challenge modeling during transitional phases, \textbf{delayed cardiorespiratory equilibrium}~\citep{xu1999oxygen} causes temporal misalignments, and \textbf{individual differences} in body composition, training status, and metabolic efficiency add significant variability. 

Despite these challenges, our optimized neural kalman based model (128-4) achieved an aggregate MAPE of 11.4\%, with most runners exhibiting errors below 20\%. This performance is particularly notable given that our method relies solely on consumer-grade wearable data and constructs the entire VO$_{2}$ trajectory from only the first second of metabolic data, without requiring exhaustive laboratory calibration. For these results, we used ground truth HR measurements rather than predicted values to isolate the performance of the VO$_{2}$ estimation component. Appendix~\ref{A.5} contains complementary results using predicted HR values from our trained models defined in previous sections (Section~\ref{Framework1}).

Figure~\ref{fig:vo2_predictions} demonstrates the generative capability of the model by plotting ground truth VO$_{2}$ measurements (blue) alongside predictions (orange) during an incremental testing protocol. The model reconstructs the entire VO$_{2}$ trajectory using only the first second of metabolic data, maintaining high fidelity over extended periods despite varying exercise intensities. Appendix~\ref{figsVO2} contains additional visualizations, including a maximal capacity assessment.

The incremental testing protocol presents a complex scenario, with 35 minutes of progressively faster 1200-meter sets until exhaustion. Despite periodic dips in the ground truth data caused by brief interruptions for blood lactate sampling, the model maintains a MAPE of 13.26\%, successfully tracking the overall increasing trend in VO$_{2}$ while effectively filtering out these transient artifacts.

\textbf{Performance Analysis and Limitations.} The 128-4 model reveals significant inter-subject variation, with MAPE ranging from 5.35\% to 21.26\%. This variation underscores the complexity of individual metabolic responses. The deeper 128-4 architecture consistently outperforms the shallower 256-2 model, suggesting that additional neural layers more effectively capture the intricate dynamics of VO$_{2}$ time-series.

With an MAE of 251\,ml/min, we achieve approximately 9\% error at 3000\,ml/min—a remarkable accomplishment using only one second of initial metabolic data. Our Kalman-inspired neural model successfully balances pointwise accuracy with realistic temporal dynamics, though challenges persist in modeling rapid high-intensity transitions and individual metabolic variability.

This sequence-to-sequence generation approach represents, to the best of our knowledge, the first method to predict complete instantaneous VO$_{2}$ trajectories using minimal initial data and smartwatch signals. Unlike existing approaches that either provide single-point VO$_{2}$max estimates or require expensive continuous respiratory measurements, our method offers a practical solution for real-world metabolic monitoring. \textbf{The performance of approximately 13\% MAPE for most participants—establishes a new benchmark in wearable-based VO$_{2}$ prediction}. Future research will explore incorporating additional physiological signals, refining temporal representations, and developing adaptive calibration methods to account for individual fitness changes over time.

\section{Discussion and Conclusion}\label{conclusion}
Our work bridges the gap between laboratory-grade physiological assessment and consumer wearables by introducing what we believe is the first framework capable of predicting instantaneous VO$_{2}$ trajectories directly from consumer devices. By combining physiologically constrained modeling with modern ML techniques, we demonstrate how sophisticated metabolic metrics—previously confined to specialized laboratories—can be reliably estimated using everyday technology, democratizing advanced exercise physiology.

\textbf{VO$_{2}$ Prediction and HR Dynamics.} Our sequence-to-sequence VO$_{2}$ prediction framework generates complete metabolic trajectories requiring only the initial second of data for calibration, achieving approximately 13\% MAPE for most participants without specialized laboratory equipment. This breakthrough enables sophisticated metabolic insights previously available only through expensive respiratory gas analysis systems costing over 30,000\$. Supporting this innovation, our HR dynamics models achieve exceptional accuracy (MAE: 2.81\,bpm, correlation: 0.87) across diverse running conditions while maintaining physiological plausibility even during high-intensity exercise (MAE: 1.99\,bpm in controlled conditions). These models not only support VO$_{2}$ prediction but also address signal interruptions in wearable PPG sensors—a persistent challenge in consumer devices.

\textbf{Toward Metabolic Zone Classification.} Our results establish a foundation for predicting metabolic thresholds from non-invasive data. Our synchronized dataset with blood lactate measurements enables future classification of aerobic, threshold, and anaerobic zones, potentially transforming training prescription by making advanced zone-based guidance accessible beyond elite athletes.

\textbf{Limitations and Future Directions.} Despite promising results, limitations include: (1) our participant cohort consists primarily of highly trained runners, potentially limiting generalizability; (2) our models do not incorporate environmental factors (temperature, humidity) that influence physiological responses; and (3) the approach does not account for longitudinal adaptations from systematic training. Future work will focus on broadening subject demographics, integrating environmental sensor data, developing longitudinal models for evolving fitness levels, and incorporating biomechanical efficiency metrics to refine VO$_{2}$ predictions.

\textbf{Safe and Responsible Innovation Statement.} Our framework for predicting instantaneous VO$_{2}$ from consumer wearables prioritizes privacy through local data processing without external transmission, while acknowledging potential bias in our high-performing athletes-focused models despite evaluated cross-runner generalizability. While democratizing metabolic monitoring offers significant health benefits, we recognize misinterpretation risks without professional guidance. Future work will incorporate broader demographics, transparent confidence metrics, and clear guidelines for responsible fitness and healthcare applications.

\newpage
\bibliographystyle{plain}
\bibliography{references}

\newpage
\newpage
\section{Appendices}\label{sup}
\appendix
\setcounter{page}{1}
This appendix provides additional implementation details, model architectures, and performance visualizations that complement the main text.

\section{Comprehensive Running Dataset Statistics}\label{RUNDATASET}
Table~\ref{tab:runner_dataset_stats} presents statistics for the HR modeling dataset, detailing the number of sessions, accumulated training time (in both seconds and minutes), and total distance covered by each participant. The dataset contaisn 831 running sessions from 20 participants, representing over 52,825 minutes (approximately 880 hours) of running activity and covering a cumulative distance of 10,222.90 kilometers. This extensive collection provides a robust foundation for developing and validating our HR prediction models across diverse runners and training conditions. Notable variance exists in individual contributions, with participant data ranging from 3 sessions (runner-5) to 124 sessions (runner-7), highlighting the dataset's capacity to capture both intra- and inter-subject variability in cardiovascular responses to running.
\begin{table}[!t]
\centering
\caption{Comprehensive running dataset statistics showing number of sessions, total time, and total distance for each participant.}
\label{tab:runner_dataset_stats}
\begin{tabularx}{\columnwidth}{l *{4}{>{\centering\arraybackslash}X}}
\hline
\multirow{2}{*}{\textbf{Runner}} & \multirow{2}{*}{\textbf{\# Sessions}} & \multicolumn{2}{c}{\textbf{Time}} & \multirow{2}{*}{\textbf{Dis. (km)}} \\ \cline{3-4}
 &  & \textbf{s} & \textbf{min} &  \\ \hline
runner-1  & 74  & 91,232  & 1,521  & 305.01  \\
runner-2  & 19  & 26,569  & 443  & 90.23  \\
runner-3  & 29  & 32,739  & 546  & 117.42  \\
runner-4  & 25  & 25,734  & 429  & 102.80  \\
runner-5  & 3  & 4,389  & 73  & 18.26  \\
runner-6  & 53  & 50,908  & 848  & 190.51  \\
runner-7  & 124  & 610,995  & 10,183  & 2,040.84  \\
runner-8  & 41  & 142,175  & 2,370  & 471.17  \\
runner-9  & 25  & 94,731  & 1,579  & 314.23  \\
runner-10  & 15  & 14,971  & 250  & 56.24  \\
runner-11  & 33  & 186,187  & 3,103  & 662.15  \\
runner-12  & 48  & 232,667  & 3,878  & 812.43  \\
runner-13  & 22  & 22,711  & 379  & 66.07  \\
runner-14  & 68  & 351,054  & 5,851  & 1,240.50  \\
runner-15  & 34  & 86,637  & 1,444  & 321.47  \\
runner-16  & 10  & 60,356  & 1,006  & 211.03  \\
runner-17  & 32  & 154,292  & 2,572  & 485.47  \\
runner-19  & 76  & 418,703  & 6,978  & 1,161.24  \\
runner-20  & 100  & 562,421  & 9,374  & 1,555.86  \\ \hline
\textbf{Total}  & \textbf{831}  & \textbf{3,169,471}  & \textbf{52,824}  & \textbf{10,222.90}  \\ \hline
\end{tabularx}
\end{table}

\section{Detailed Mathematical Formulation of HR Models, VO$_{2}$ Model and Training Methodology}
We provide detailed mathematical formulations for our two HR prediction approaches: the ODE-based model and the Kalman filter model. Both models share a common feature processing pipeline but diverge in how they model cardiac dynamics.
\subsection{ODE-based Heart Rate Model}\label{HRModelAppendix}
We model HR as a nonlinear dynamical system whose evolution depends on
workout-specific inputs and remains within physiologically plausible
bounds.  Learnable maps are denoted $f_{\bullet}$ with parameters
$\bm{\vartheta}_{\bullet}$.

\paragraph{Inputs and latent backbone.}
Given a multivariate input window
$\mathbf{x}_{t}\!\in\!\mathbb{R}^{B\times T\times D_{\text{in}}}$ (batch
size $B$, length $T$, feature dimension $D_{\text{in}}$), a feature
encoder produces latent embeddings
\begin{equation}
  \mathbf{e}_{t}=f_{\text{enc}}\!\bigl(\mathbf{x}_{t};\bm{\vartheta}_{\text{enc}}\bigr),\nonumber
\end{equation}
followed by a recurrent backbone
\begin{equation}
  \mathbf{s}_{t}=f_{\text{temp}}\!\bigl(\mathbf{e}_{t},\mathbf{s}_{t-T};
        \bm{\vartheta}_{\text{temp}}\bigr),\qquad
  \mathbf{s}_{0}=\mathbf{0},\nonumber
\end{equation}
that captures long-range temporal context.

\paragraph{Latent ODE dynamics.}
We introduce a latent HR state
$g_{t}\!\in\!\mathbb{R}^{B\times T\times D_{\text{latent}}}$ governed by
a first-order ODE
\begin{equation}
  \frac{\partial g_{t}}{\partial t}=f_{\text{ODE}}
       \bigl(g_{t},\mathbf{s}_{t};\bm{\vartheta}_{\text{ODE}}\bigr),\nonumber
\end{equation}
whose right-hand side embodies cardiac demand
\begin{equation}
  f_{\text{ODE}}\!\bigl(g_{t},\mathbf{s}_{t};\bm{\vartheta}_{\text{ODE}}\bigr)
        = d(\mathbf{s}_{t};\bm{\vartheta}_{\text{dec}}) - g_{t},
  \nonumber
\end{equation}
with $d(\cdot)$ a small MLP that estimates the HR level required by the
current exercise intensity.

\paragraph{Numerical integration and learning.}
The ODE is integrated with a fourth-order Runge–Kutta solver (RK4), and
gradients are obtained by the adjoint method of Neural ODEs, enabling
end-to-end training.

\paragraph{Physiological normalisation and decoding.}
From the backbone state we predict per-batch normalisation parameters
\begin{align}
  \hat{\mu}_{t} &= f_{\mu}(\mathbf{s}_{t};\bm{\vartheta}_{\text{nrm}})
      \in\mathbb{R}^{B}, \nonumber\\
  \hat{\sigma}_{t} &= f_{\sigma}(\mathbf{s}_{t};\bm{\vartheta}_{\text{nrm}})
      \in\mathbb{R}^{B}, \nonumber
\end{align}
then decode observable HR by an affine map
\begin{equation}
  \hat{h}_{t}=f_{\text{dec}}\!\bigl(g_{t},\mathbf{s}_{t};
                \bm{\vartheta}_{\text{dec}}\bigr)
            = \hat{\sigma}_{t}\,g_{t}+\hat{\mu}_{t}. \nonumber
\end{equation}

\paragraph{Interpretability and advantages.}
Our ODE model offers several advantages for HR prediction: (1) Bounded parameters enforce physiological plausibility; (2) The ODE captures gradual HR adaptation to changing intensity; (3) The model remains interpretable, with parameters corresponding to meaningful physiological quantities.

\subsection{Kalman Filter Heart Rate Model}
\label{KalmanAppendix}
We cast HR prediction as a state-estimation problem solved by an enhanced Kalman filter whose parameters are supplied by neural networks.
As before, learnable maps are denoted $f_{\bullet}$ with
parameters $\bm{\vartheta}_{\bullet}$.

\paragraph{Inputs and latent backbone.}
A multivariate input window
$\mathbf{x}_{t}\!\in\!\mathbb{R}^{B\times T\times D_{\text{in}}}$ is
encoded
\[
\mathbf{e}_{t}=f_{\text{enc}}(\mathbf{x}_{t};\bm{\vartheta}_{\text{enc}}),
\qquad
\mathbf{s}_{t}=f_{\text{temp}}(\mathbf{e}_{t},\mathbf{s}_{t-T};
        \bm{\vartheta}_{\text{temp}}),\;
\mathbf{s}_{0}=0,
\]
yielding a workout embedding $\mathbf{s}_{t}\!\in\!\mathbb{R}^{B\times
D_{\text{hid}}}$ that conditions all filter parameters.

\paragraph{State definition.}
We track latent HR and its velocity via the two-dimensional state
\[
\mathbf{z}_{t}
   =\begin{bmatrix}g_{t}\\\dot g_{t}\end{bmatrix}
   \in\mathbb{R}^{B\times2},\qquad
\mathbf{P}_{t}\in\mathbb{R}^{B\times2\times2}
\]
with process noise covariance
$\mathbf{Q}_{t}=f_{\text{noise}}(\mathbf{s}_{t};\bm{\vartheta}_{\text{noise}})$
and measurement noise variance
$\mathbf{R}_{t}=f_{\text{meas}}(\mathbf{s}_{t};\bm{\vartheta}_{\text{meas}})$.

\paragraph{Kalman recursion.}
\emph{Prediction step}
\begin{align}
\mathbf{z}_{t}^{+} &=\mathbf{z}_{t}
     +f_{\text{tr}}(\mathbf{s}_{t},\mathbf{z}_{t};\bm{\vartheta}_{\text{tr}}), \nonumber
   \label{eq:kal_pred_state}\\
\mathbf{P}_{t}^{+} &=\mathbf{P}_{t}+\mathbf{Q}_{t}.\nonumber
\end{align}

\emph{Update step}  
\begin{align}
\hat{h}_{t}  &=f_{\text{obs}} (\mathbf{s}_{t};\bm{\vartheta}_{\text{obs}}),\nonumber\\
\nu_{t}      &=\hat{h}_{t}-g_{t}^{+},\nonumber\\
k_{t} &= f_{\text{gain}}\!\bigl(\mathbf{P}_{t}^{+},\mathbf{R}_{t};
                                \bm{\vartheta}_{\text{gain}}\bigr),
\quad\text{where }\mathbf{P}_{t}^{+}=\mathbf{P}_{t}+\mathbf{Q}_{t}.\nonumber
\end{align}
State and covariance are corrected by the scalar gain $k_{t}$:
\begin{align}
g_{t+1}      &=g_{t}^{+}+k_{t}\nu_{t},\nonumber\\
\dot g_{t+1} &= \dot g_{t}^{+} + \gamma_t\,k_t\,\nu_t,
\quad\text{with }\gamma_t = 0.5,\nonumber\\
\mathbf{P}_{t+1}&=
   \mathbf{P}_{t}^{+}\odot
   \begin{bmatrix}
     1-k_{t} & 0\\
     0       & 1-\gamma_t k_t
   \end{bmatrix}.\nonumber
\end{align}
The fixed $0.5$ factor damps velocity corrections, yielding smoother
trajectories.

\paragraph{Physiological bounds.}
After each update we clip to a biologically plausible range:
\[
g_{t}\;\leftarrow\;
\min\bigl(\text{HR}_{\max},\;\max(\text{HR}_{\min},\,g_{t})\bigr).
\]

\paragraph{Network architecture.}
All auxiliary maps $f_{\text{enc}}$, $f_{\text{tr}}$, $f_{\text{obs}}$, $f_{\text{noise}}$,
 $f_{\text{meas}}$ and $f_{\text{gain}}$ are small MLPs, while $f_{\text{temp}}$ is a GRU.

\paragraph{Advantages.}
Our Kalman filter approach offers several advantages: (1) The two-dimensional state vector captures both HR and its rate of change, enabling better tracking of cardiac dynamics; (2) The adaptive Kalman gain responds to prediction errors, increasing when errors are high to give more weight to new measurements; (3) Neural network prediction of filter parameters allows adaptation to different exercise contexts; (4) The reduced gain for velocity updates creates smoother trajectories while maintaining responsiveness to intensity changes.

\subsection{Heart Rate Models Training and Implementation}\label{Appex:TrainingHR}
Both models were implemented in PyTorch and trained using similar procedures. We trained two GRU backbones that differ only in width and depth:
\begin{center}
\begin{tabular}{@{}lcc@{}}
\toprule

Variant & Hidden dim. & GRU layers \\ \midrule
\textsc{Large} & 128 & 2 \\
\textsc{Small} & 64  & 3 \\ \bottomrule
\end{tabular}
\end{center}
All feed‑forward heads (encoders, decoders, gain/demand networks, etc.) are two‑layer MLPs with LeakyReLU activations and dropout 0.1. We optimized the model parameters using the Adam optimizer (weight‑decay $10^{-5}$, gradient clipping $\lVert g\rVert_2\le 1.0$) with an initial learning rate of 0.001, reducing the learning rate by a factor of 0.5 when validation loss plateaued for 10 epochs. To prevent overfitting, we employed early stopping with a patience of 20 epochs. For the \textbf{Kalman} model we used \texttt{ReduceLROnPlateau} with \texttt{factor}=0.5 and \texttt{patience}=10, and an extended early‑stopping window of 100 epochs. The \textbf{ODE} model employed the same scheduler but with \texttt{factor}=0.2, a cooldown of 3 epochs, a minimum
learning rate of $10^{-6}$, and the 20‑epoch patience already noted above. We trained with a batch size of 32 and the default Adam momentum parameters \(\beta_1=0.9,\;\beta_2=0.999\).

For both models, we used a masked mean absolute error (MAE) loss function to handle variable-length sequences:
\begin{equation}
    \mathcal{L}_{\text{MAE}} =
  \frac{\sum_{b=1}^B \sum_{t=1}^T
        |h_{b,t}^{\text{pred}} - h_{b,t}^{\text{true}}| \cdot \text{mask}_{b,t}}
       {\sum_{b=1}^B \sum_{t=1}^T \text{mask}_{b,t}}
\end{equation}

Additionally, we incorporated an auxiliary loss to encourage accurate parameter prediction:
\begin{equation}
\mathcal{L}_{\text{aux}} = ||\hat{\mu} - \mu_{\text{obs}}||_1 + ||\hat{\sigma} - \sigma_{\text{obs}}||_1
\end{equation}
where $\mu_{\text{obs}}$ and $\sigma_{\text{obs}}$ are the observed mean and standard deviation of HR in each sequence, and $\hat{\mu}$ and $\hat{\sigma}$ are the predicted parameters.

The total loss was a weighted combination:
\begin{equation}
\mathcal{L}_{\text{total}} = \mathcal{L}_{\text{MAE}} + \lambda_{\text{aux}} \cdot \mathcal{L}_{\text{aux}}
\end{equation}
with $\lambda_{\text{aux}} = 0.1$.

Both models achieved comparable performance, with the ODE model excelling in capturing long-term physiological trends and the Kalman filter model showing advantages in responding to rapid intensity changes. The models' complementary strengths suggest potential benefits in ensemble approaches for future work.

\subsection{Kalman-based VO$_{2}$ Model}\label{KalmanVO2Appendix}
We frame oxygen‑consumption dynamics as a state–estimation problem with physiologically informed constraints.  
Let $v_t\!\in\!\mathbb{R}$ denote the latent VO$_2$ state at time $t$, with $\hat{v}_t$ its filtered estimate and $\tilde{v}_{t^-}$ the trend‑extrapolated prediction.  
As throughout the paper, learnable functions are written $f(\,\cdot\,;\bm{\vartheta})$, where the subscript on~$\bm{\vartheta}$ indicates the module.

\paragraph{Temporal feature extraction.}
The network processes a window $\mathbf{x}_t\in\mathbb{R}^{B\times T\times D_{\text{in}}}$  
(containing biomechanical features, HR, and temporal context):
\begin{equation}
\mathbf{h}_t
= f_{\text{temp}}\!\bigl(f_{\text{enc}}(\mathbf{x}_t;\bm{\vartheta}_{\text{enc}})\,;\bm{\vartheta}_{\text{temp}}\bigr),\nonumber
\end{equation}
where $f_{\text{temp}}$ is a bidirectional GRU and
$\mathbf{h}_t\!\in\!\mathbb{R}^{B\times D_h}$ is the hidden representation at time $t$.

\paragraph{Process and measurement-noise prediction.}
Neural heads produce time-varying covariance surrogates from the hidden state:
\begin{align}
\mathbf{P}_{\text{process},t}
   &= f_{\text{process}}(\mathbf{h}_t;\bm{\vartheta}_{\text{process}}),\nonumber\\
\mathbf{R}_{\text{meas},t}
   &= f_{\text{measurement}}(\mathbf{h}_t;\bm{\vartheta}_{\text{measurement}}).\nonumber
\end{align}

\paragraph{Neural–Kalman update.}
Given $\mathbf{h}_t$, the VO$_2$ estimate is updated via
\begin{equation}
\hat{v}_{t}^{\text{KF}}= \hat{v}_{t-1}
+ f_{\text{gain}}(\mathbf{h}_t;\bm{\vartheta}_{\text{KF}})
\bigl[f_{\text{obs}}(\mathbf{h}_t;\bm{\vartheta}_{\text{obs}})-\hat{v}_{t-1}\bigr]_{%
      f_{\Delta_{\min}}(\mathbf{h}_t;\bm{\vartheta}_{\Delta})}^{%
      f_{\Delta_{\max}}(\mathbf{h}_t;\bm{\vartheta}_{\Delta})}\,,\nonumber
\end{equation}
where the adaptive gain $f_{\text{gain}}$ and clamping limits
$f_{\Delta_{\min}},f_{\Delta_{\max}}$ are themselves MLPs, allowing
data‑driven confidence weighting and physiologically plausible
rate‑of‑change bounds.

\paragraph{Trend/direct dual pathway.}
To improve robustness,
\begin{align}
\tilde{v}_{t^-} &=
   \hat{v}_{t-1}
   + f_{\text{trend}}(\mathbf{h}_{t};\bm{\vartheta}_{\text{trend}})
     \bigl(\hat{v}_{t-1}-\hat{v}_{t-2}\bigr),\nonumber\\
\hat{v}_{t}^{\text{dir}} &= f_{\text{direct}}(\mathbf{h}_{t};\bm{\vartheta}_{\text{dir}}),\nonumber\\
\hat{v}_{t} &= \beta_{t} \hat{v}_{t}^{\text{dir}} + (1-\beta_{t}) \hat{v}_{t}^{\text{KF}},\nonumber
\end{align}

\paragraph{Implementation details.}
All heads  
$f_{\text{enc}}$, $f_{\text{gain}}$, $f_{\text{obs}}$,
$f_{\Delta_{\min}}$, $f_{\Delta_{\max}}$,
$f_{\text{process}}$, $f_{\text{measurement}}$,
$f_{\text{trend}}$, $f_{\text{blend}}$, and $f_{\text{direct}}$
are two-layer MLPs with LeakyReLU activations and dropout 0.1, parameterized by
$\bm{\vartheta}_{\text{enc}}$, $\bm{\vartheta}_{\text{KF}}$, $\bm{\vartheta}_{\text{obs}}$, $\bm{\vartheta}_{\Delta}$, $\bm{\vartheta}_{\text{trend}}$, $\bm{\vartheta}_{\text{blend}}$, and $\bm{\vartheta}_{\text{dir}}$, respectively.
The temporal module $f_{\text{temp}}$ is a biGRU
(two layers, hidden size $D_h=128$).  
All outputs are trained end‑to‑end with an MSE loss on breath‑by‑breath
VO$_2$ targets; clamping reduces gradient explosion when rapid
transients exceed physiological bounds.

\subsection{Training and Implementation Details}
\label{trainingVO2Appendix}
We employ the composite loss (main text) with base, dynamic, and auxiliary components:
\begin{align}
\mathcal{L} &= 
  \underbrace{\mathbb{E}_t\bigl|\hat{v}_t - v_t\bigr|}_{\text{MAE}}
+ \lambda_{\text{dynamic}} 
  \underbrace{\mathbb{E}_t\bigl(\alpha\bigl|\tfrac{d\hat{v}_t}{dt}-\tfrac{dv_t}{dt}\bigr|
    +\beta\bigl|\tfrac{d^2\hat{v}_t}{dt^2}-\tfrac{d^2v_t}{dt^2}\bigr|\bigr)}_{\text{derivative loss}} \nonumber\\
&\quad+ \lambda_{\text{aux}}
  \underbrace{\mathbb{E}_t\bigl(|\hat{\mu}_t-\mu_t| + |\hat{\sigma}_t-\sigma_t|
    + |\hat{\Delta}_t-\Delta_t|\bigr)}_{\text{moment loss}},\nonumber
\end{align}

The dynamic loss captures temporal patterns using both first and second derivatives:
\begin{align}
\mathbb{E}_t \Bigl( \alpha \Bigl| \frac{d\hat{v}_t}{dt} - \frac{dv_t}{dt} \Bigr|
  + \beta \Bigl| \frac{d^2\hat{v}_t}{dt^2} - \frac{d^2v_t}{dt^2} \Bigr| \Bigr).\nonumber
\end{align}

The auxiliary loss enforces statistical consistency between predicted and observed distributions:
\begin{align}
\mathbb{E}_t | \hat{\mu}_t - \mu_t | &= |\hat{\mu} - \mu| \nonumber \\ 
\mathbb{E}_t | \hat{\sigma}_t - \sigma_t | &= |\hat{\sigma} - \sigma| \nonumber \\ 
\mathbb{E}_t | \hat{\Delta}_t - \Delta_t | &= |\hat{\Delta} - \Delta|, \nonumber
\end{align} where $\mu_t$ is the temporal mean, $\sigma_t$ is the standard deviation, and $\Delta_t = Q_{0.95}(|v_\tau - v_{\tau-1}|)$ represents the 95th percentile of absolute differences in consecutive VO$_2$ values.

\paragraph{Curriculum weights.}  
At epoch $e$:  
$\lambda_{\text{base}}=\max(0.30,\,1-e/20)$,  
$\lambda_{\text{dynamic}}=1-\lambda_{\text{base}}$,  
$\lambda_{\text{aux}}=\min(0.30,\,0.10+0.01e)$.

\paragraph{Model variants.}
\begin{center}
\begin{tabular}{@{}lccc@{}}
\toprule
Variant & GRU hidden & GRU layers & Head MLP layers \\ \midrule
Small   & 128 & 2 & 2 \\
Large   & 256 & 4 & 4 \\ \bottomrule
\end{tabular}
\end{center}

\paragraph{Optimisation.}
AdamW (\text{lr} =\,$4\times10^{-3}$, weight‑decay $10^{-5}$)  
with cosine‑annealing warm restarts (T$_0$ = 10, T$_\text{mult}$=2, $\eta_{\min}=10^{-6}$).  
Batch 32, sequence length 60s (1Hz); gradient clipping $\|g\|_2\le 1+4e^{-e/10}$;  
early stopping after 50 epochs without validation‑MAE improvement.

\section{Additional Results}
\subsection{Additional HR Figures}\label{HRADDITIONALFIGURES}
\begin{figure*}[p!]
\centering
\caption{Ten random heart rate prediction using the \textit{Standard} predictions mode on random sampled sessions with true HR (blue) and predicted HR (red). Each subfigure shows a single session’s timeline, along with mean absolute error (MAE), root mean square error (RMSE), mean absolute percentage error (MAPE), and correlation.}
\Description{Ten representative HR prediction sessions using the standard mode, showing true (blue) and predicted (red) heart rate traces. Each subplot displays workout timeline with performance metrics including Mean Absolute Error (MAE), Root Mean Square Error (RMSE), Mean Absolute Percentage Error (MAPE), and correlation coefficient, demonstrating the model's accuracy across diverse running conditions and exercise intensities.}
  \includegraphics[scale=0.40]{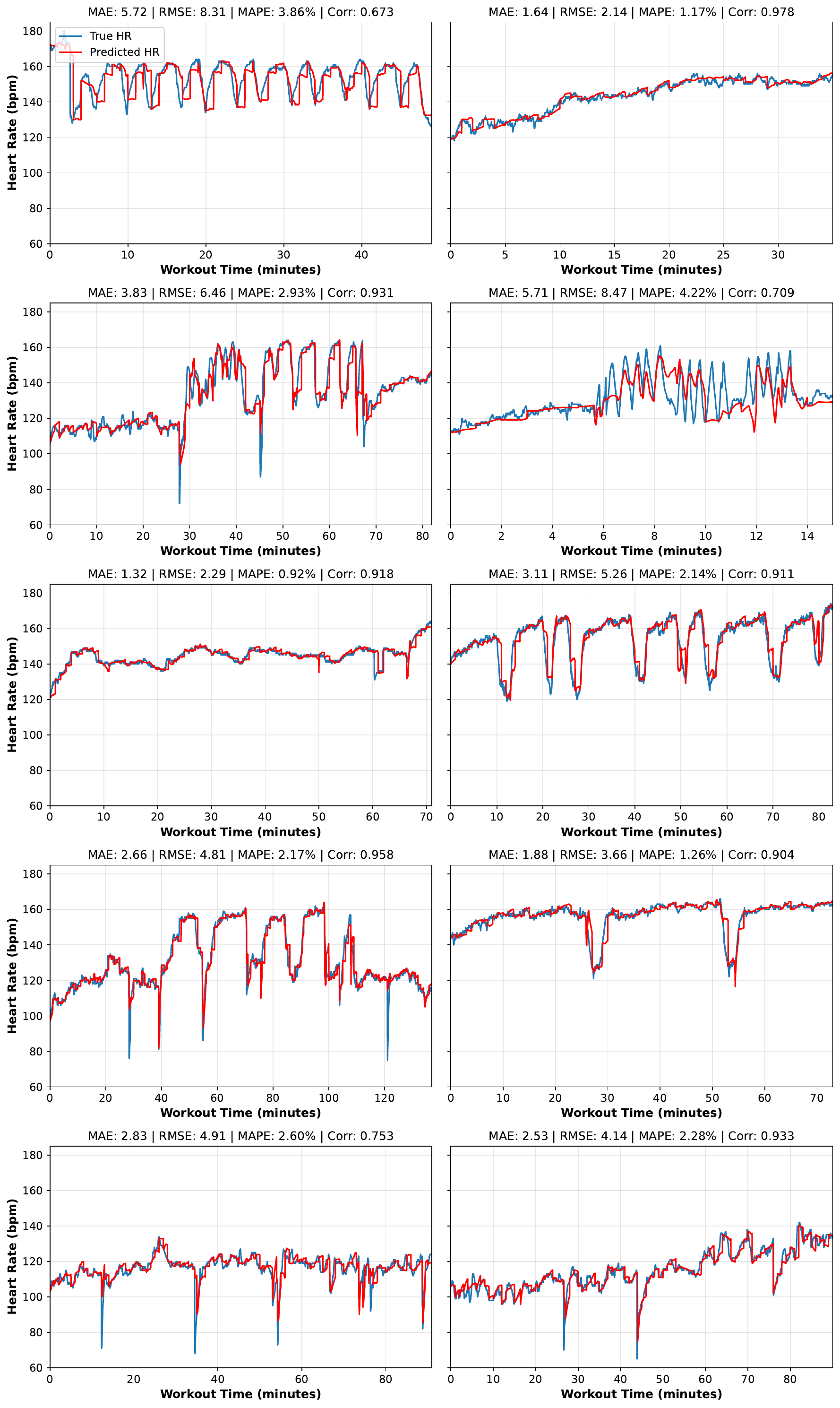}
\label{fig:all_sessions}
\end{figure*}

\begin{figure*}[p!]
\centering
\caption{Ten random heart rate prediction using the \textit{Generating} predictions mode on random sampled sessions with true HR (blue) and predicted HR (red). Each subfigure highlights a unique session with corresponding performance metrics (MAE, RMSE, MAPE, and correlation).}
\Description{Ten representative HR prediction sessions using the Generating mode, showing true (blue) and predicted (red) heart rate traces. Each subplot displays workout timeline with performance metrics including Mean Absolute Error (MAE), Root Mean Square Error (RMSE), Mean Absolute Percentage Error (MAPE), and correlation coefficient, demonstrating the model's accuracy across diverse running conditions and exercise intensities.}
\includegraphics[scale=0.40]{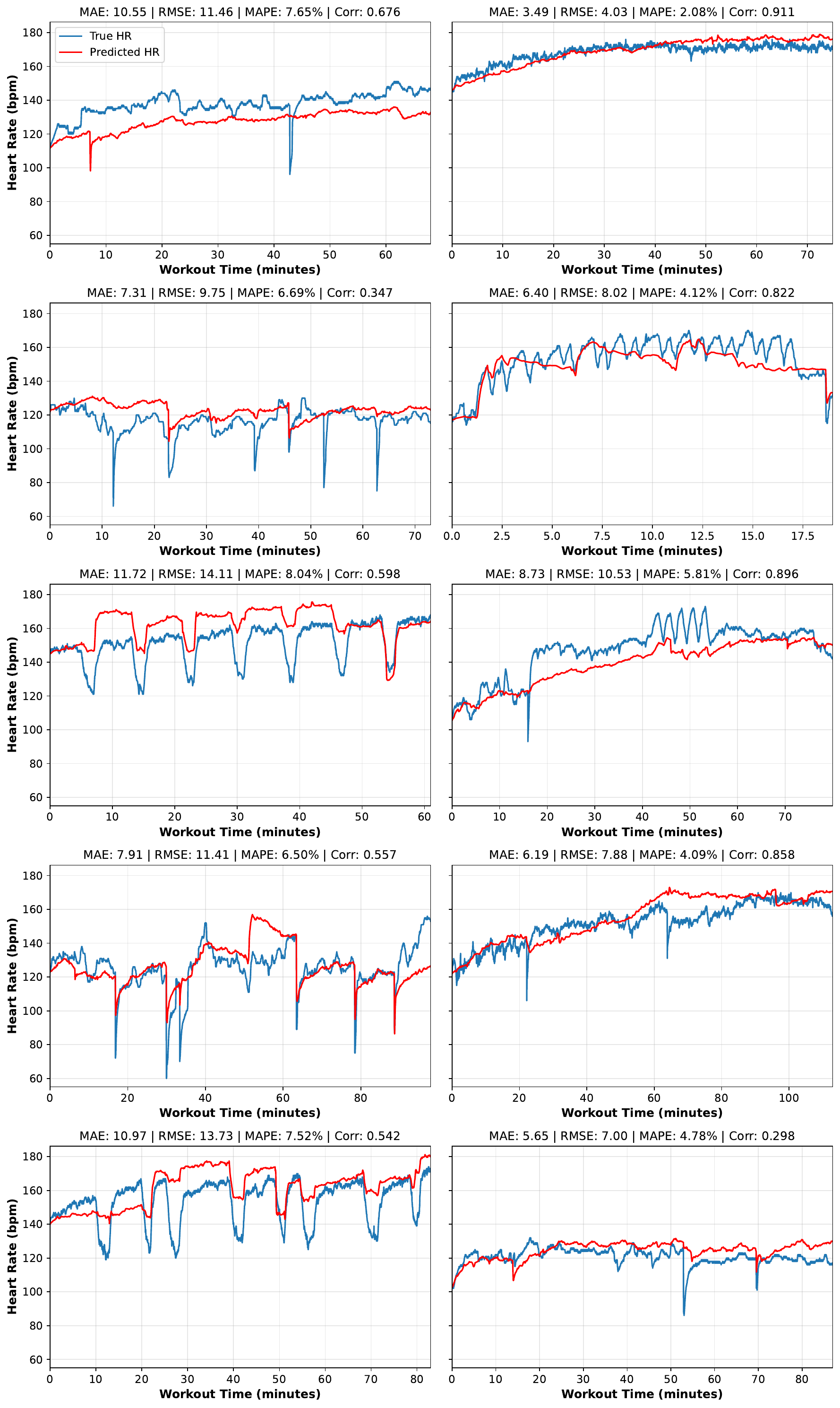}
\label{fig:all_ten_sessions}
\end{figure*}

Figure~\ref{fig:all_sessions} presents additional HR prediction using the standard prediction mode, while Figure~\ref{fig:all_ten_sessions} presents using the the generative mode.
\clearpage

\subsection{Additional VO$_{2}$ results using Two Trained HR Models and figures}\label{A.5}
\begin{table}[t]
\centering
\caption{Comparison of Sequence-to-Sequence Instantaneous VO$_{2}$ Prediction for Models 256-2 and 128-4 using Leave-One-Out Cross-Validation. HR used is the predicted HR values using generation mode. Bold values indicate the better (lower) performance.}
\begin{tabularx}{\columnwidth}{l *{6}{>{\centering\arraybackslash}X}}
\toprule
& \multicolumn{3}{c}{KalmanVO$_2$ 256-2} & \multicolumn{3}{c}{KalmanVO$_2$ 128-4} \\
\cmidrule(r){2-4} \cmidrule(l){5-7}
Runner & MAE & RMSE & MAPE & MAE & RMSE & MAPE \\
& & & (\%) & & & (\%) \\
\midrule
\multicolumn{7}{l}{\textit{With ODE-based HR predictions (128-2)}} \\
\midrule
Runner-1 & 645.78 & 682.96 & 19.24 & \textbf{519.27} & \textbf{565.93} & \textbf{15.74} \\
Runner-2 & 1326.20 & 1361.69 & 35.40 & \textbf{463.79} & \textbf{518.90} & \textbf{12.94} \\
Runner-3 & 250.61 & 481.78 & 12.36 & 234.11 & 475.18 & \textbf{12.04} \\
Runner-4 & 287.74 & 335.78 & 11.01 & \textbf{128.56} & \textbf{182.10} & \textbf{6.03} \\
Runner-5 & 697.68 & 717.53 & 25.03 & \textbf{549.20} & \textbf{571.91} & \textbf{19.88} \\
Runner-6 & 347.87 & 450.75 & 15.61 & 402.03 & 495.38 & 17.04 \\
Runner-7 & 193.00 & 219.82 & 7.16 & \textbf{122.36} & \textbf{160.31} & \textbf{4.70} \\
Runner-8 & 321.80 & 382.35 & 13.54 & 266.80 & 347.77 & \textbf{11.98} \\
Runner-9 & 333.82 & 407.08 & 25.83 & 268.31 & 344.64 & \textbf{21.32} \\
Runner-10 & 407.44 & 514.14 & 14.41 & 398.53 & 491.12 & \textbf{14.23} \\
\midrule
Aggregate & 481.19 & 555.39 & 17.96 & \textbf{335.30} & \textbf{415.32} & \textbf{13.59} \\
\midrule
\multicolumn{7}{l}{\textit{With Kalman-based HR predictions (128-2)}} \\
\midrule
Runner-1 & 747.67 & 781.70 & 21.97 & \textbf{529.75} & \textbf{574.35} & \textbf{16.03} \\
Runner-2 & 1326.20 & 1361.69 & 35.40 & 1757.21 & 1804.52 & 46.63 \\
Runner-3 & 262.15 & 479.42 & 12.59 & 290.56 & 495.09 & 13.59 \\
Runner-4 & 439.11 & 500.90 & 15.78 & \textbf{137.36} & \textbf{189.68} & \textbf{6.30} \\
Runner-5 & 697.75 & 717.59 & 25.03 & \textbf{549.20} & \textbf{571.91} & \textbf{19.88} \\
Runner-6 & 383.54 & 473.57 & 16.73 & 463.29 & 550.49 & 18.92 \\
Runner-7 & 192.96 & 219.79 & 7.15 & \textbf{120.33} & \textbf{158.70} & \textbf{4.63} \\
Runner-8 & 321.80 & 382.35 & 13.54 & 288.13 & 345.58 & \textbf{12.21} \\
Runner-9 & 396.71 & 449.44 & 29.59 & 274.75 & 354.99 & \textbf{21.83} \\
Runner-10 & 468.62 & 559.96 & 16.36 & 432.30 & 529.91 & \textbf{15.29} \\
\midrule
Aggregate & 523.65 & 592.64 & 19.41 & 484.29 & 557.52 & 17.53 \\
\bottomrule
\end{tabularx}
\label{VO2PRedHR}
\end{table}
To quantify how HR–prediction quality propagates into downstream metabolic estimation, we trained two sequence-to-sequence VO$_2$ models that share the same architecture except for backbone size (\textbf{256-2} versus \textbf{128-4}; see Table~\ref{VO2PRedHR}. Both networks were evaluated in leave-one-runner-out cross-validation using HR signals that were \emph{predicted} in generation mode by either our best ODE-based or Kalman-based HR model (both 128-2). The table below reports per-runner and aggregate MAE, RMSE, and MAPE. Boldface highlights the better of the two VO$_2$ variants under each HR source. The table shows that the results are worse when using the predicted HR compared to when using the true HR (see in the section), although the results are still well below the 20\% MAPE threshold which is the considered standard. 

\begin{figure*}[p!]
\centering
\caption{Sequence-to-Sequence VO$_2$ Prediction Across Different Sessions using true HR values.}
\Description{Representative VO$_2$ prediction sessions.}
  \includegraphics[scale=0.52]{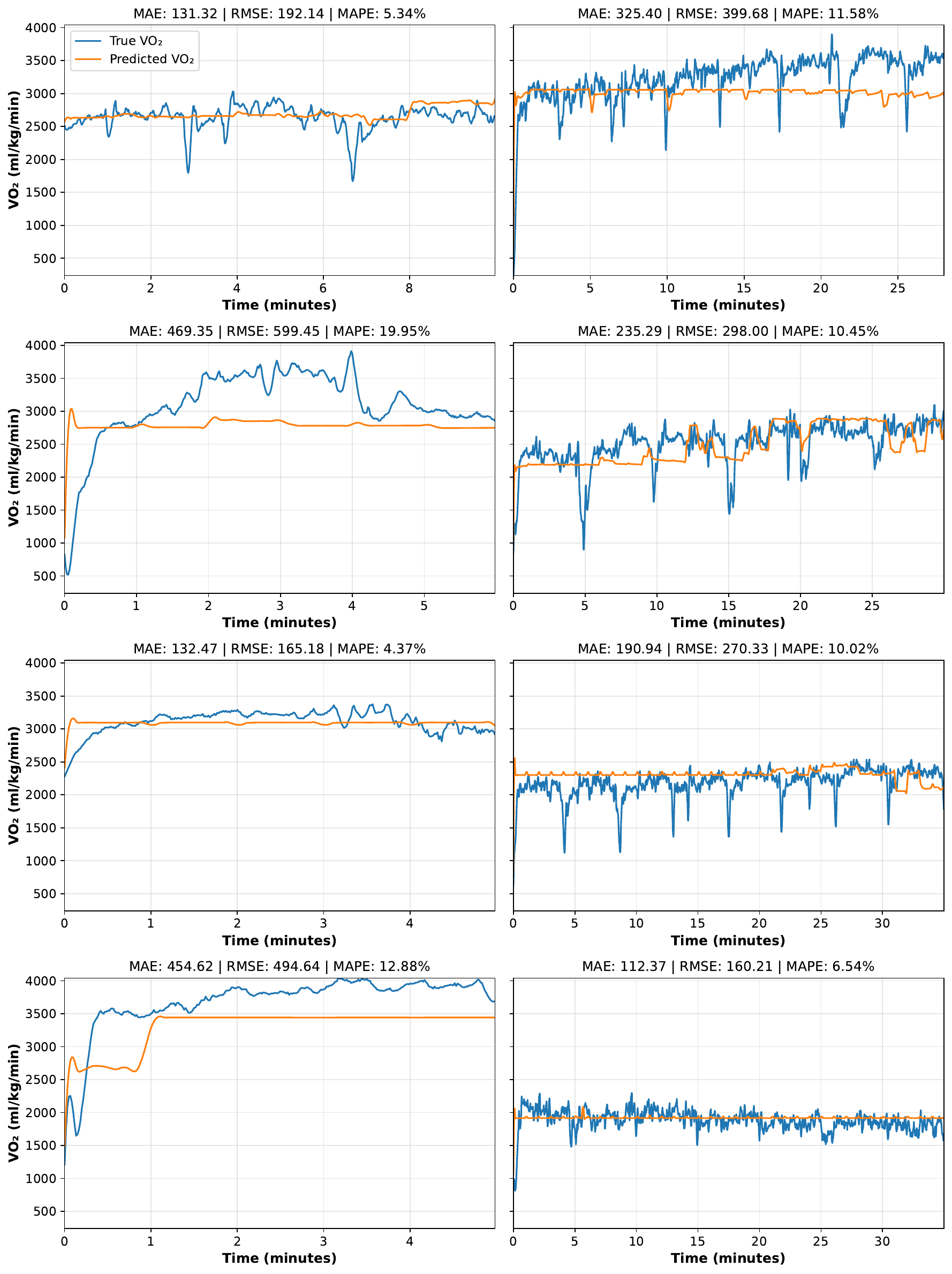}
\label{fig:vo2_predictions_comprehensive}
\end{figure*}

\subsection{Additional Figures for Predicting VO$_{2}$}\label{figsVO2}
Figure~\ref{fig:vo2_predictions_comprehensive} presents additional VO$_{2}$ prediction visualizations.

\end{document}